# Foundation Model–Driven Semantic Change Detection in Remote Sensing Imagery

Hengtong Shen, Li Yan, Hong Xie, Yaxuan Wei, Xinhao Li, Wenfei Shen, Peixian Lv and Fei Tan

***Abstract*—Remote sensing (RS) change detection is essential for interpreting surface dynamics. Semantic change detection (SCD) further enables pixel-level understanding of multi-class transitions, yet it remains highly sensitive to pseudo-changes induced by imaging conditions. Recent RS foundation models exhibit strong representation capabilities, and their ability to extract semantically consistent features across different temporal instances and environmental conditions is particularly critical for mitigating pseudo-changes in SCD tasks. However, due to variations in model architectures and the challenges of adapting to multi-scale features, existing SCD methods are typically designed as fixed and rigid frameworks tailored to specific backbone networks, lacking the modular flexibility required to seamlessly accommodate the diverse and hierarchical multi-scale feature representations produced by emerging RS foundation models. To address these adaptation challenges, we first introduce a modular Cascaded Gated Decoder (CG-Decoder) designed to serve as a flexible bridge between various backbones and SCD tasks. The CG-Decoder processes multi-scale features in a coarse-to-fine manner while promoting effective multi-level feature interaction and adaptive change extraction. Building upon this, we present PerASCD, a robust and unified SCD framework driven by the RS foundation model PerA. Furthermore, we propose a Soft Semantic Consistency Loss (SSCLoss) to mitigate the numerical instability often encountered during the training of large-scale models. Extensive experiments on the SECOND and LandsatSCD benchmarks demonstrate that PerASCD achieves new state-of-the-art SeK scores (26.11% and 65.21%), surpassing the previous best method by 0.61% and 4.95%, respectively, while also exhibiting favorable data efficiency, seamless cross-backbone generalization, and enhanced interpretability. Beyond numerical superiority, our approach demonstrates robust semantic consistency in the face of radiometric variations, providing a more interpretable and reliable solution for SCD. The code is available at https://github.com/SathShen/PerASCD.git.**

Manuscript received xx xx, 2026; revised xx xx, 2026; accepted xx xx 2026. Date of publication xx xx 2026; date of current version xx xx 2026. This work was partially supported by the National Key R&D Program of China (Grant No. 2025YFB3910302), the Guangxi Key R&D Program (Grant No. GKFN2600640051), the Zhejiang Province "Vanguard" and "Geese Leading" Research and Development Plan (Grant No. 2025C01073), and the National Natural Science Foundation of China (Grant No. 42394061). *(Hengtong Shen and Li Yan are co-first authors.) (Corresponding author: Hong Xie.)*

The authors are with the School of Geodesy and Geomatics, Wuhan University, Wuhan 430079, China (e-mail: shenht@whu.edu.cn; lyan@sgg.whu.edu.cn; hxie@sgg.whu.edu.cn; weiyaxuan@whu.edu.cn; lxh17866703382@163.com;15505084554@163.com; lvpeixian@whu.edu.cn; 2024282140087@whu.edu.cn).

## I. INTRODUCTION

SEMANTIC change detection (SCD) or multi-class change detection have emerged in recent years as a RS interpretation task that has been receiving increasing attention. Unlike conventional binary change detection (BCD), SCD does not merely identify *where* changes have occurred, it further specifies the *from-to* semantics of those changes (i.e., the class transition). This capability enables the extraction of semantic-level change maps from RS imagery, supporting more accurate urban planning [1], natural resource management [2], and disaster assessment [3].

Binary change detection (BCD) extracts change information and focus on detecting the locations of changed pixels. Due to the disregard for pixels' categories, BCD often fails to characterize the semantic change information required in downstream applications. SCD networks can not only predict the precise locations of change areas but also provide detailed land-cover class information across multiple temporal images. However, the challenges of multi-class annotation, the complexity of semantic change tasks, and the relatively low demand for semantic change tasks, most existing change detection studies have focused on BCD. With advances in deep learning and RS imaging technologies, researchers are no longer satisfied with merely locating changes. Instead, there is a growing need to identify the specific types of change to support more precise decision-making across diverse domains.

In recent years, numerous SCD algorithms based on deep neural networks have been proposed, which have achieved impressive progress and played an important role in multiple fields. Simultaneously, the rapid progress in RS foundation models has provided a promising direction for advancing this field. Foundation models are large-scale deep learning models pre-trained on vast and diverse datasets, possessing powerful general representation capabilities that can be adapted to multiple downstream tasks [4], [5], [6],. In particular, their ability to extract semantically consistent features across different temporal instances is critical for capturing complex "from-to" transitions in RS imagery.

However, although RS foundation models offer superior semantic understanding, their potential in SCD remains largely underexplored. Current SCD algorithms still suffer from various deficiencies ranging from model design to task logic, which significantly hinder the effective integration and performance of high-capacity backbones. In general, these limitations can be summarized as follows:

(1) Sensitivity to pseudo-changes: Most current SCD methods utilize lightweight encoders pre-trained on natural images, which often struggle to distinguish land-cover transitions from pseudo-changes (e.g., seasonal variations or illumination shifts) prevalent in complex RS scenarios. While emerging RS foundation models offer the potential to extract semantically consistent features across different temporal instances to mitigate these effects, their application in SCD remains largely underexplored.

(2) Lack of structural adaptability: Existing SCD decoders are predominantly designed as fixed, rigid architectures tailored for specific backbones. They lack the modular flexibility required to seamlessly adapt to the diverse and hierarchical multi-scale feature representations produced by various backbones. This rigidity leads to insufficient feature interaction and restricts the generalizability of SCD frameworks across different vision encoders.

(3) Training instability and category imbalance: The inherent category imbalance in SCD tasks, where unchanged regions predominate, significantly exacerbates training difficulties and predisposes models to rely on simplistic "shortcut" solutions. Although auxiliary constraints like Semantic Consistency Loss (SCLoss) have been introduced to align semantic and change representations, they frequently encounter numerical instability under specific conditions. This instability hinders the stable convergence and reliable deployment of high-capacity models in RS interpretation.

How to bridge the representation gap between RS foundation models and SCD, enabling effective utilization of temporally consistent and multi-scale semantic features for robust change understanding?

To address these challenges, we propose PerASCD, a robust and unified framework designed to effectively bridge various RS foundation models with the SCD task. Our approach prioritizes structural adaptability, training effectiveness, and stability rather than mere architectural complexity. Specifically, we leverage the self-supervised pre-trained foundation model PerA [7] to extract temporally invariant semantic representations, which serve as the foundation for suppressing pseudo-changes in complex landscapes. To overcome the rigidity of conventional SCD pipelines, we introduce the Cascaded Gated Decoder (CG-Decoder). As a modular and flexible component, the CG-Decoder utilizes a Change-Aware Gating Module (CAGM) to adaptively modulate feature weights across multiple scales. This mechanism acts as a dynamic semantic refiner, facilitating a coarse-to-fine information propagation process that ensures precise change localization while filtering out irrelevant background noise. Furthermore, we provide a deep dive into the training dynamics of SCD models, introducing the Soft Semantic Consistency Loss (SSCLoss) to resolve the numerical instability issues encountered when training SCD models with SCLoss. The major contributions of this work are summarized as follows:

(1) We propose PerASCD, which systematically integrates RS foundation models into the SCD task. By leveraging the semantic consistency of large-scale pre-trained encoders, our framework significantly enhances the model’s robustness against pseudo-changes, achieving state-of-the-art (SOTA) performance and favorable data efficiency—outperforming the full-data baseline with only 50% of the training data.

(2) We introduce the modular Cascaded Gated Decoder (CG-Decoder) to address the structural rigidity of existing SCD methods. Its coarse-to-fine multi-scale processing and Change-Aware Gating Module (CAGM) enable adaptive feature arbitration, providing the flexibility to accommodate diverse hierarchical features. This design not only allows seamless adaptation across various vision encoders, but also enhances interpretability by explicitly localizing and suppressing pseudo-changes.

(3) We provide a comprehensive analysis of the numerical instability associated with SCLoss and propose SSCLoss to improve training stability under challenging optimization settings. Experiments on LandsatSCD demonstrate that SSCLoss mitigates the observed instability under mixed-precision and high-learning-rate training, providing a practical strategy for stable optimization of large-scale SCD models.

## II. Related Work

In this section, we review the current state of RS foundation models and the development of BCD and SCD methods. We particularly focus on the evolution of SCD architectural paradigms.

### *A. RS Foundation Model*

Foundation models are general-purpose models pre-trained on large-scale, diverse datasets. RS foundation models, pre-trained on vast RS data and tailored to its unique characteristics, outperform general models pre-trained on natural images in RS interpretation tasks, demonstrating superior generalization and domain-specific capabilities. Recent advances in self-supervised learning have produced numerous RS foundation models to address high annotation costs. Generative approaches include RingMo [5] and Scale-MAE [8], while contrastive learning-based methods such as SeCo [9] and SkySense [10] have gained prominence due to their effectiveness and flexibility.

The core mechanism behind the success of these models lies in their ability to learn spatio-temporal semantic consistency [11], [12]. Unlike conventional encoders, recent RS foundation models are pre-trained to capture temporally invariant features in RS images by reconstructing structural essence (in Masked Image Modeling) or aligning multi-view representations (in contrastive learning) across diverse environmental conditions. This contextual reasoning forces the encoder to prioritize high-level geometric and semantic structures over low-level pixel fluctuations. Consequently, these models exhibit a natural resistance to pseudo-changes, such as shifts in solar angles, atmospheric conditions, and seasonal vegetation variations, which typically plague lightweight models. These studies demonstrate the strong potential of self-supervised pre-training for RS tasks. SCD demands such robust semantic understanding to accurately capture genuine "from-to" transitions, making RS foundation models a highly promising direction for advancing SCD.

### *B. Binary Change Detection*

Early change detection methods were predominantly focused on binary change detection (BCD). Traditional BCD approaches relied heavily on handcrafted features and pixel-wise classifiers to generate binary change maps, suffering from limited accuracy and poor generalization capability. With the advent of deep Convolutional Neural Networks (CNNs) and Transformers, deep learning-based methods have gradually replaced conventional techniques, achieving substantial performance improvements in BCD tasks. Most deep learning-based methods for BCD employ siamese-like network architectures to extract features from bi-temporal images, followed by difference comparison and feature fusion to predict binary change regions. In the early stages, methods such as FC-Siam-conc [13] and SNUNet-CD [14] utilized purely CNN structures for BCD tasks. Following the widespread adoption of ViTs [15], ViT-based or CNN-ViT hybrid approaches like BIT [16], ChangeFormer [17], have gained prominence. In recent works on RS foundation models [5], [10], has been commonly employed as a key downstream task for model evaluation. However, in the contrast, SCD has received scarcely any attention in this context.

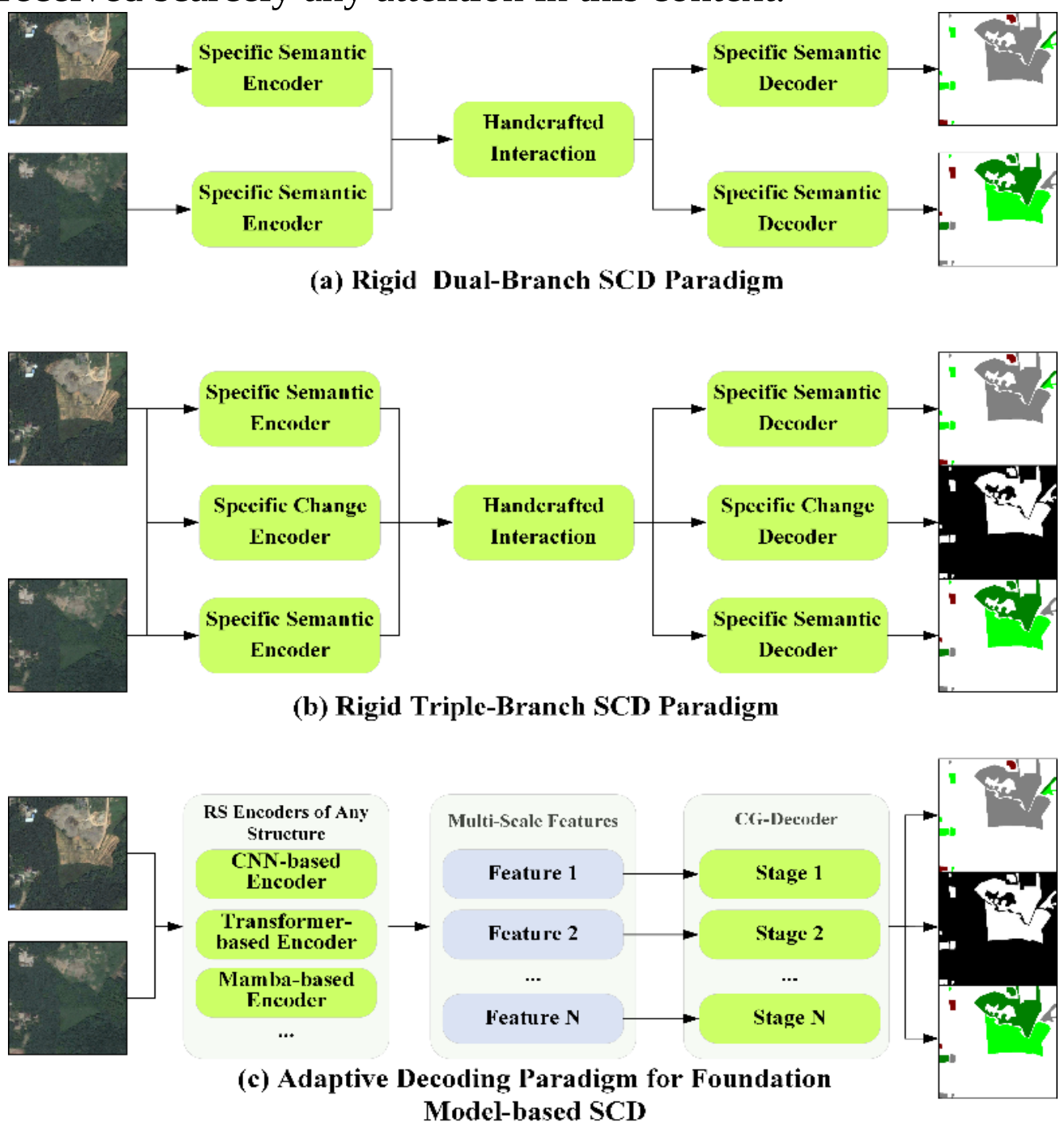

**Fig. 1.** Existing common paradigms of semantic change detection models: (a) Rigid Dual-Branch: A fixed architecture which change masks are derived from independent semantic branches; (b) Rigid Triple-Branch: A fixed architecture with an explicit change-specific decoder; (c) Adaptive Integration (Ours): A backbone-agnostic paradigm that adaptively fuses multi-scale features from foundation models.

### *C. Semantic Change Detection*

Early explorations in pixel-level SCD were predominantly either non-end-to-end approaches [18] or methods that only output post-change semantic pixel-level classifications [19], [20]. HRSCD [21] is widely regarded as the pioneering effort that aligns with the contemporary definition of SCD, establishing the foundation for "from-to" change outputs. It systematically investigated the feasibility of various network architectures in this context. Subsequently, multiple end-to-end SCD methods have been continually proposed. Existing SCD paradigms primarily employ siamese-like dual-temporal encoders. However, they have evolved from early rigid designs toward the adaptive integration of foundation models proposed in this work.

As illustrated in Fig. 1(a), methods like DCCANet [22] and SCDNet [23] closely approximate the basic logic of SCD by extracting bi-temporal features for independent semantic segmentation, and then generating change masks directly from these semantic maps. However, because they lack explicit supervision for change-specific features, their change detection capability is entirely bottlenecked by the independent accuracy of the two semantic branches, often leading to inferior performance. As illustrated in Fig. 1(b), to address the shortcomings of dual-branch designs, triple-branch methods—such as SemanticCD [24], MCDnet [25], GCF-SCD-Net [26], and SCanNet [27]—have gained widespread popularity. The superiority of this paradigm stems from the introduction of a separate, explicit decoder specifically designed to process change information. This dedicated branch provides direct constraints and facilitates collaborative interaction between semantic identification and change localization, consistently yielding better performance than dual-branch architectures. Despite their progress, the aforementioned methods are constrained by fixed, rigid decoding structures that are meticulously tailored to specific lightweight backbones. They suffer from excessive manual design and inflexible handcrafted interaction mechanisms. More importantly, emerging RS foundation models produce feature representations with significantly higher dimensions and deeper semantics. Traditional fixed decoders lack the structural modularity required to handle these rich representations, making it exceedingly difficult for them to achieve the efficient adaptation of multi-scale features.

To overcome the aforementioned adaptation barriers, Fig. 1(c) illustrates the adaptive foundation model-based SCD paradigm proposed in this work. Unlike the rigid couplings in (a) and (b), our paradigm treats the encoder as a backbone-agnostic component, capable of accommodating diverse RS foundation model architecture, ranging from CNN-based and Transformer-based to the emerging Mamba-based encoders. These high-capacity backbones produce a hierarchy of multi-scale features (Feature 1 to $N$) with varying spatial resolutions and semantic depths. To efficiently ingest these complex representations, the CG-Decoder is designed as a modular, cascaded interface. Through successive Stages (Stage 1 to $N$), it adaptively aligns and fuses features from different levels, replacing the redundant multi-branch piping of legacy methods with a streamlined, unified flow. This design ensures a seamless interface to unlock the full potential of diverse RS foundation models, facilitating precise change localization while significantly reducing the need for manual architectural re-engineering.

Auxiliary information from other modalities has also been explored to enhance performance. For instance, MSCD-Net [28] incorporates SAR imagery and DSM data to aid semantic understanding, demonstrating significant advantages. RB-SCD [29] incorporates text and frequency domain information into the network, providing a new approach to SCD methods. Furthermore, several recent works like Mamba-FCS [30], FAPMNet [31] and GSTM-SCD [32] employ new encoder architectures to improve efficiency and performance, providing new insights for SCD model designing. While these methods enhance the feature extraction capabilities of the models, the RS-specific semantic consistency provided by these encoders remains limited. We leverage a RS foundation model and augment the utility and adaptability of the features during the decoding stage.

Furthermore, beyond architectural paradigms and model designs, addressing the severe category imbalance remains a fundamental challenge in SCD, as the dominance of unchanged regions often biases the expected gradient within a mini-batch [21], [33]. These non-change regions contain numerous pseudo-changes caused by environmental variations, which require stronger semantic consistency to suppress[34], [35]. Meanwhile, because targets in remote sensing imagery range from small-scale changes (e.g., vehicles, rooftops) to large-scale transformations (e.g., urban expansion), multi-scale features must be progressively refined in a coarse-to-fine manner to systematically eliminate semantic ambiguities across scales and accurately localize genuine change regions. Consequently, an effective SCD decoder should function not merely as a feature aggregator, but as a multi-scale confidence-based feature regulator. This motivation drives our design of the Cascaded Gated Decoder (CG-Decoder), which processes change information through successive stages to accommodate these multi-scale variations. Within this framework, the Change-Aware Gating Module (CAGM) is introduced to implicitly learn this adaptive regulation, prioritizing gating resources to suppress confusing non-change objects without relying on rigid and counterproductive explicit supervision. Finally, to ensure the reliability of this complex optimization process, we address the numerical instability inherent in existing consistency constraints like SCLoss [36] by proposing the Soft Semantic Consistency Loss (SSCLoss), which provides a more stable training trajectory for high-capacity SCD models. We discussed this in detail in Section III-D.

## III. Method

In this section, we first briefly introduce the pre-trained model and fine-tuning strategy adopted in this work. We then present the overall framework of PerASCD. Subsequently, the

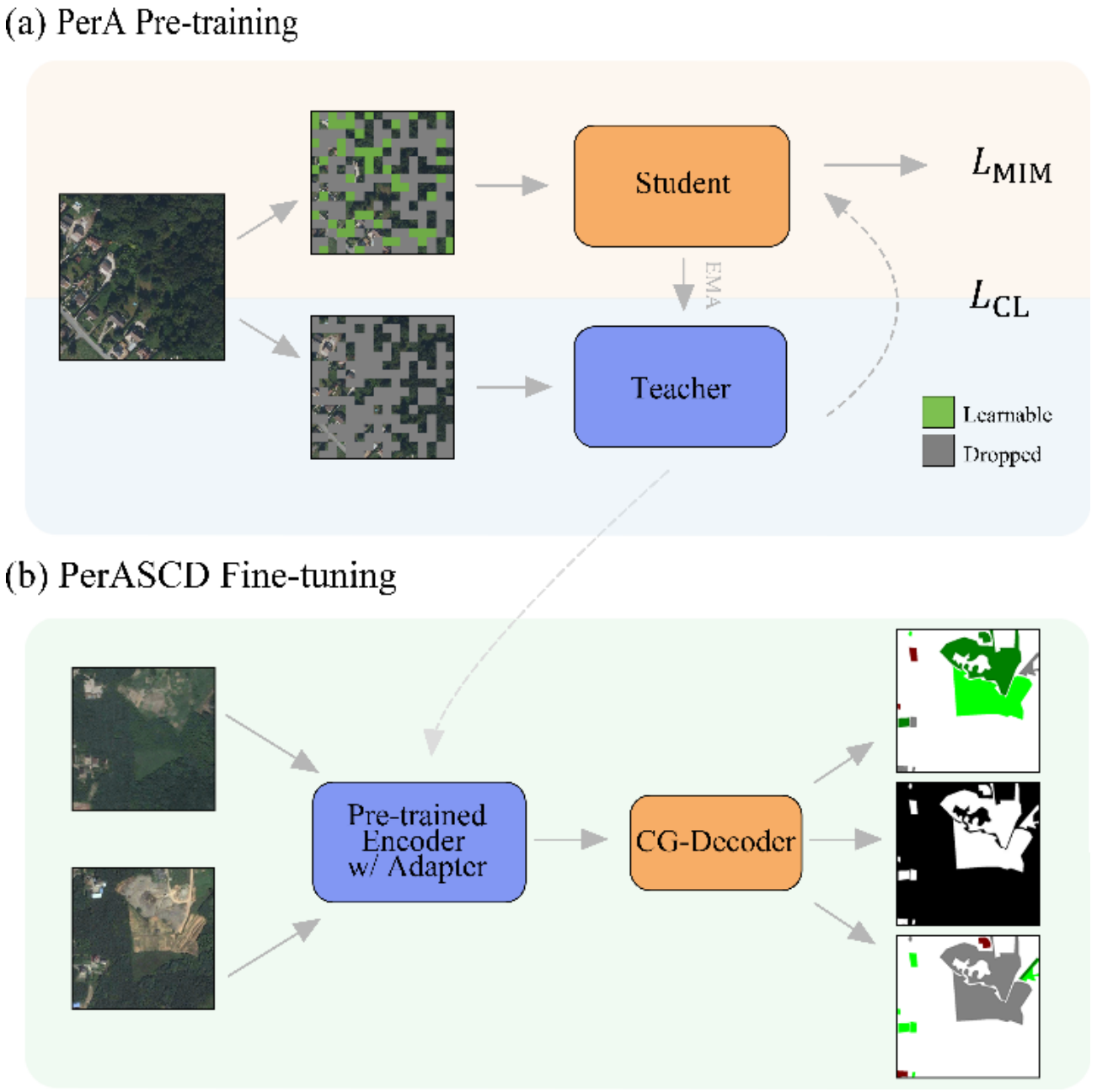

**Fig. 2.** Overall pipeline: (a) PerA Pre-training: A self-supervised paradigm combining contrastive learning and MIM, with the teacher encoder extracted for downstream initialization; (b) PerASCD Fine-tuning: The pre-trained ViT backbone, enhanced by a ViT-Adapter, extracts multi-scale features. These are refined by the CG-Decoder to generate bi-temporal semantic maps and a binary change mask.

Cascaded Gated Decoder and Change-Aware Gating Module are described in detail. Finally, we introduce the Soft Semantic Consistency Loss.

### *A. Overall pipeline*

As shown in Fig. 2(a), the PerA pre-training method is a self-supervised learning approach that integrates the strengths of contrastive learning and masked image modeling. Specifically, PerA replaces random cropping in the standard contrastive pretext task of contrastive learning with random masking and incorporates learnable mask tokens into the input patches. This design enables the model to acquire global semantic understanding from RS images via contrastive objectives while simultaneously capturing fine-grained local semantic details through pixel-level reconstruction, significantly enhancing the inherent capability of the ViT model for semantic feature extraction and comprehension in RS imagery. Pre-trained on a million-scale high-resolution RS dataset, PerA encourages the encoder to learn invariant semantic representations rather than low-level radiometric details. This characteristic is highly consistent with the requirements of SCD, where the model should suppress pseudo-changes caused by illumination, seasonal variation, and imaging differences, while remaining sensitive to genuine “from-to” semantic transitions. Therefore, PerA provides a suitable foundation backbone for extracting temporally consistent and spatially detailed representations in our PerASCD framework. After pre-training, we employ the teacher encoder of PerA, a standard ViT as the pre-trained

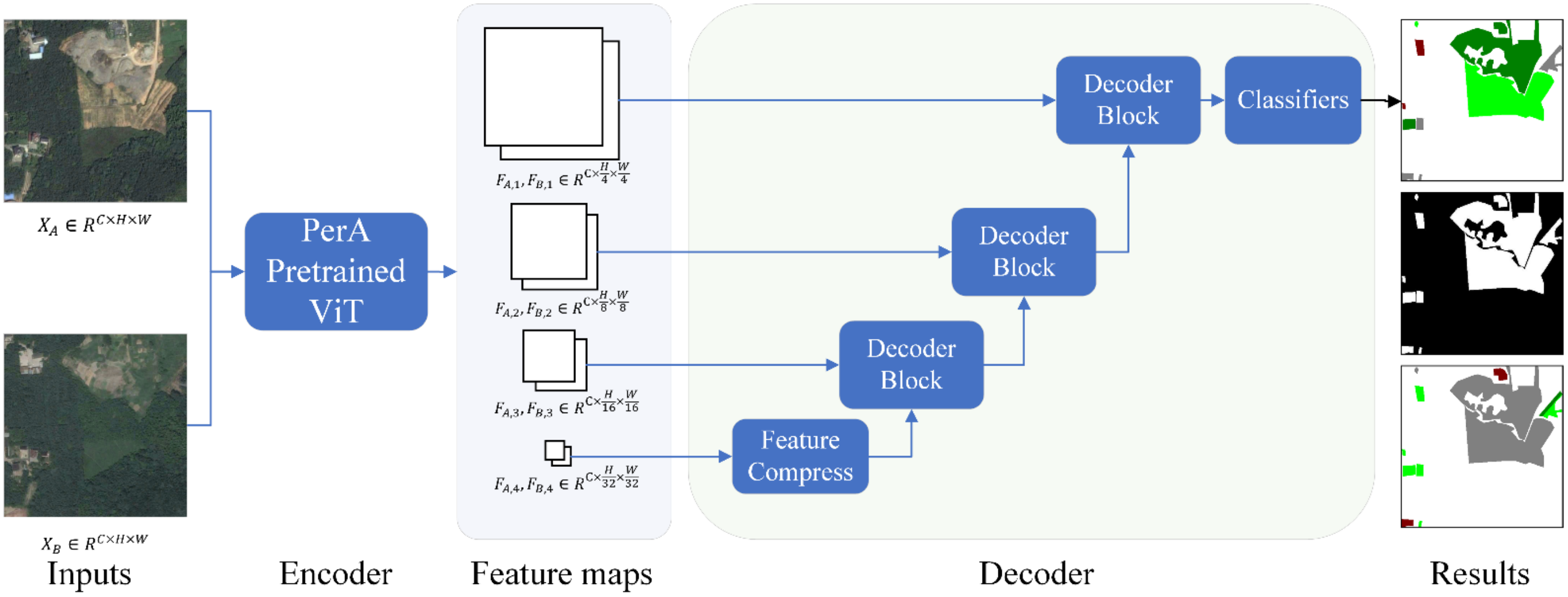

**Fig. 3.** Architecture of PerASCD.

backbone for downstream transfer learning.

As illustrated in Fig. 2(b), after the pre-training stage, the pretrained encoder is adopted as the backbone of PerASCD and combined with the CG-Decoder for fine-tuning on the SCD task. We employ ViT-Adapter [37] to enhance the pretrained model with stronger spatial priors, improving pixel-level prediction accuracy and enabling the extraction of multi-scale features. The multi-scale features generated by the encoder are progressively processed, upsampled, and fused by the CG-Decoder, ultimately producing both the bi-temporal semantic change map and the binary change map. Detailed descriptions of PerASCD and the adopted CG-Decoder are provided in the subsequent sections.

### *B. Architecture of PerASCD*

The overview of proposed PerASCD architecture is shown in Fig. 3. PerASCD adopts a simple encoder-decoder architecture and leverages the PerA method which is self-supervised pre-trained via contrastive learning on massive unlabeled RS image tiles. PerASCD first extracts multi-scale, semantically feature maps using the PerA pre-trained ViT encoder and ViT-Adapter. Given the bi-temporal RS images $X_A, X_B \in R^{C\times H\times W}$, the pre-trained model processes them into multi-scale feature maps $F_{A,1}, F_{B,1}, \dots, F_{A,4}, F_{B,4}$.These features are then progressively processed through CG-Decoder blocks. This cascaded multi-scale feature processing paradigm draws inspiration from U-Net's skip connections [38]. We will provide a detailed description of this in Section III-C. Through this design, the CG-Decoder achieves efficient feature compression, multi-level semantic fusion, precise change extraction, and multi-scale upsampling within a simple, modular framework. Finally, the features are processed and fused into a unified set of features, which are then fed into the classifier to produce the final binary change map and the semantic maps.

### *C. Cascaded Gated Decoder and Change-Aware Gating Module*

The Cascade-Gated Decoder (CG-Decoder) represents the core innovation in PerASCD, enabling effective semantic change decoding of multi-scale bi-temporal features. Through high-dimensional feature compression, multi-level feature fusion, feature upsampling, and change-aware gating, it transforms complex multi-scale features into fused pixel-level semantic representations suitable for change detection, achieving seamless multi-scale bridging. Specifically, this mechanism acts as a dynamic semantic refiner that facilitates a coarse-to-fine information propagation process, ensuring precise change localization while filtering out irrelevant pseudo-changes. This decoder simultaneously addresses feature dimensionality reduction, fusion, and upsampling while accommodating diverse encoder input requirements, simplifying and streamlining the SCD task. The structure of a single CG-Decoder Block is illustrated in Fig. 4(a).

The Cascade-Gated Decoder is composed of multiple CG-Decoder Blocks, facilitating modular multi-scale feature fusion and change information extraction. Specifically, the CG-Decoder block $n$ first processes smaller-scale deep features from previous block $x_{A,n}$, $x_{B,n}$, and $x_{C,n}$ representing pre-change, post-change, and change features, respectively. The two temporal features are passed through a dual-layer convolutional network for processing. Subsequently, the three features are concatenated to integrate semantic and change information, and then fed into another dual-layer convolution to yield a more refined change representation. Following this, the three features undergo upsampling to align with the shallow features' resolution. This process can be formulated as:

$$h_A, h_B = DoubleConv(x_{A,n}, x_{B,n}) \tag{1}$$

$$h_C = DoubleConv\left(Cat(x_{C,n}, h_A, h_B)\right) \tag{2}$$

$$h'_A, h'_B, h'_C = Upsample(h_A, h_B, h_C) \tag{3}$$

where $h'_A$, $h'_B$, and $h'_C$ are the upsampled features, and $Cat$

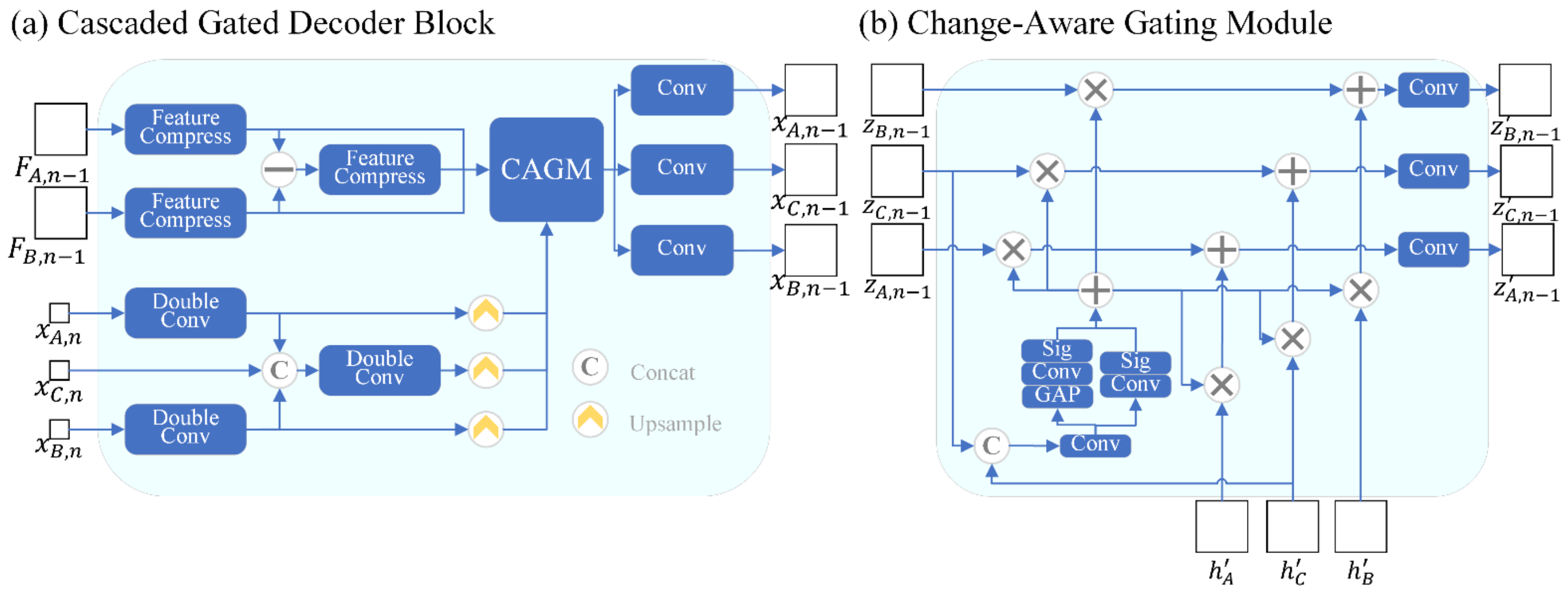

**Fig. 4.** (a) Architecture of Cascaded Gated Decoder Block; (b) Architecture of Change-Aware Gating Module.

denotes the vector concatenation operation. Larger-scale shallow features are directly sourced from the encoder outputs. Their processing mirrors that of deep features, but replaces the dual-layer convolution with a feature compress layer and the concatenation with pixel-wise differences between the pre- and post-change features. The initial deep features $x_{A,N}$, $x_{B,N}$, and $x_{C,N}$ in first block are processed similarly. This can be expressed as:

$$z_{A,n-1}, z_{B,n-1} = FC(F_{A,n-1}, F_{B,n-1}) \quad (4)$$

$$z_{C,n-1} = FC\left(abs(z_{A,n-1} - z_{B,n-1})\right) \quad (5)$$

where $z_{A,n-1}$, $z_{B,n-1}$, and $z_{C,n-1}$ are the compressed features, $FC$ denotes the feature compression layer, and $abs$ represents the absolute value operation. The FC layer employs the CBAM alongside a single-layer convolution for feature compression, minimizing information loss under minimal computational overhead. This design ensures that both high-dimensional features from large-scale foundation models and low-level features from lightweight decoders can be flexibly mapped into a unified feature space, enabling efficient adaptation across multi-dimensional and multi-scale representations, that is difficult to achieve with conventional rigid decoders. When high-dimensional features $f$ are input into FC layer, the process be expressed as:

$$f' = CBAM(f) \quad (6)$$

$$f'' = ConvCompress(f') \quad (7)$$

where $ConvCompress$ is a convolutional layer that maps the input features to the desired output dimension. The $f'$ denotes the features processed through the CBAM attention mechanism to minimize information loss, while $f''$ represents the compressed features.

The processed shallow and deep features are both fed into the Change-Aware Gating Module (CAGM) for fusion. The structure of the Change-Aware Gating Module is illustrated in Fig. 4(b). The CAGM first concatenates the input change features and generates global and local weighted maps with convolution layers. These representations are then transformed into probabilities $W_{global}$ and $W_{local}$ via Sigmoid function. We calculate final weights as $(1 + W_{global}) \odot W_{local}$, which are then applied to the corresponding shallow and deep features to the module. Subsequently, the weighted features are added together in a pixel-wise, and output to the final fusion convolution. This process can be succinctly summarized by the following formulation:

$$\mathrm{W}_{local} = \sigma\left(Conv\left(Cat(h'_c, z_{C,n-1})\right)\right) \quad (8)$$

$$\mathrm{W}_{global} = \sigma\left(Conv\left(GAP\left(Cat(h'_c, z_{C,n-1})\right)\right)\right) \quad (9)$$

$$[W_z, W_h] = Spilt((1 + W_{global}) \odot W_{local}) \quad (10)$$

$$z'_{i,n-1} = Conv(W_z \odot z_{i,n-1} + W_h \odot h'_i) \quad (11)$$

where $i \in \{A, B, C\}$, $\sigma$ denotes Sigmoid function, $GAP$ denotes global average pooling, and $Spilt$ denotes the channel-wise splitting operation that separates the generated weights into $W_z$ and $W_h$.

After the CAGM, each branch is further processed by an independent convolutional refinement layer:

$$x_{i,n-1} = Conv(z'_{i,n-1}). \quad (12)$$

These convolutional layers are used to enhance local spatial consistency, which serve as inputs to the next CG-Decoder block. and project the fused features into the feature space required by the next CG-Decoder block, yielding $x_{A,n-1}$, $x_{B,n-1}$, and $x_{C,n-1}$.

Finally, after passing through multiple CG-Decoder blocks, the bi-temporal semantic features and the change features are respectively fed into the classifier to compute probabilities, yielding the prediction results.

### *D. Soft Semantic Consistency Loss*

The Semantic Consistency Loss (SCLoss) is designed to align the semantic and change representations, and has been experimentally validated to substantially mitigate the class imbalance problem, enhancing the overall performance of SCD models. Given two feature vectors $x_1$ and $x_2$, the formula for SCLoss can be expressed as:

$$L_{SC} = \begin{cases} 1 - \cos(x_1, x_2), & if\ y = 1 \\ max(0, \cos(x_1, x_2) - m), & if\ y = -1 \end{cases} \quad (13)$$

where $cos$ denotes the cosine similarity and $m$ denotes a constant margin which is usually set to 0.1. The $y$ denotes the binary change label, with $y = 1$ indicating no change and $y = -1$ indicating a change.

However, we observe that the optimization process may exhibit training instability under certain conditions. While other factors, such as learning rate and mixed-precision training, may also contribute, the instability is observed to be substantially mitigated upon disabling the original SCLoss. Therefore, we hypothesize that these issues are due to a limitation in the mathematical formulation of SCLoss. Specifically, owing to the introduction of the margin in the SCLoss, when the feature distribution undergoes drastic changes, the cosine similarity values near the margin boundary may abruptly cross the threshold. Consequently, the gradient with respect to the feature representations can suddenly vanish or switch from zero to a non-zero value, leading to severe loss fluctuations or unstable gradient behavior during training. This issue becomes more pronounced when a large learning rate is adopted, especially in conjunction with mixed-precision training. In contrast to the commonly observed scenario in which an excessively large learning rate induces persistent oscillations or consistently low accuracy throughout training, the phenomenon under discussion manifests as an otherwise normal training process. When numerical instability occurs, however, it triggers a precipitous, cliff-like drop in accuracy.

As illustrated in Fig. 5, such issue manifests as cliff-like dips in the accuracy curves during the training process. Furthermore, through monitoring, we can clearly observe that a large proportion of pixel-wise cosine similarities exceed the margin sharply when instability events occur. Accordingly, the $F_{scd}$ values on validation set exhibit a sudden degradation corresponding to pronounced fluctuations in the loss curve.

To this end, we design a novel loss function called Soft Semantic Consistency Loss (SSCLoss) to mitigate the numerical instability while preserving the original representation alignment capability. The formulation of SSCLoss can be expressed as:

$$L_{SSC} = \begin{cases} 1 - \cos(x_1, x_2), & if\ y = 1 \\ \tau Softplus\left(\frac{\cos(x_1, x_2) - m}{\tau}\right), & if\ y = -1 \end{cases} \quad (14)$$

where:

$$Softplus = \log(1 + \exp(x)). \quad (15)$$

The $Softplus$ replace the hard hinge margin with a smooth approximation, leading to continuous gradient transitions. The temperature parameter $\tau$ jointly controls the width of the transition region around the margin and the amount of gradient leakage inside the margin, by rescaling both the argument and the magnitude of the softplus function. We empirically adjust $\tau$ for different dataset to ensure a smooth transition of gradient activation around the margin while preserving the discriminative role of the margin. This design preserves the acceptable region for feature similarity constraints defined in the original loss function, while preventing large-scale gradient population migration, significantly mitigating the likelihood of numerical instability. We will discuss this in detail in the ablation experiments of Section IV.

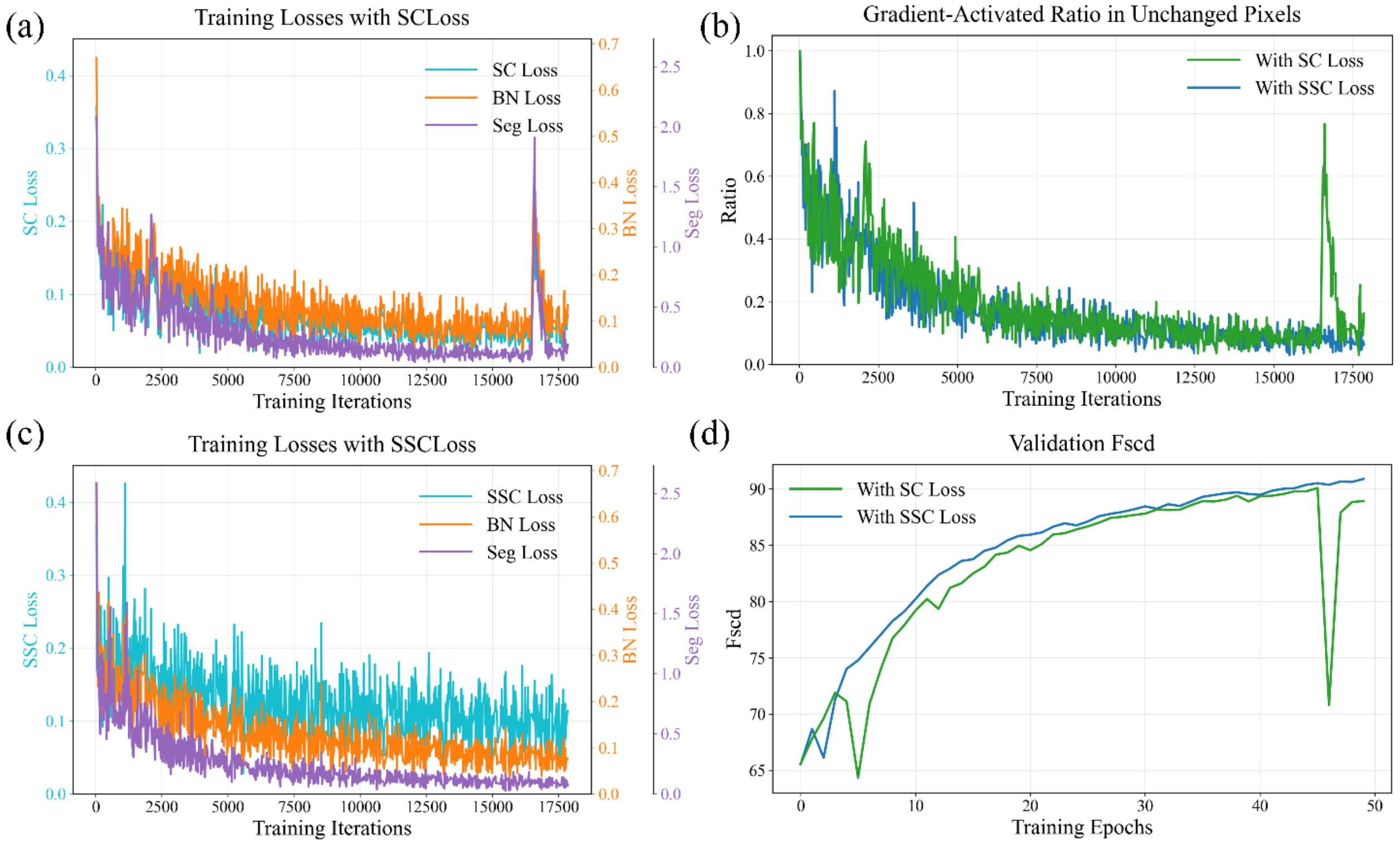

**Fig. 5.** Training curves under numerical instability (a) Training losses with SCLoss; (b) Gradient-activated ratio in unchanged pixels with different loss; (c) Training losses with SSCLoss; (d) Validation $F_{scd}$ with different loss.

## IV. Experiments

### A. Datasets

We evaluate our method on the SEmantic Change detectiON Dataset (SECOND) [39] and LandsatSCD [40]. which are publicly available benchmark datasets for SCD.

The SECOND is consist of 4662 pairs of aerial images in a high-resolution from several platforms and sensors. These pairs of images are distributed over the cities, such as Hangzhou, Chengdu, and Shanghai. Each image has been cropped into 512 × 512 size and is annotated at the pixel level. The images are in a spatial resolution varies from 0.5 to 3 m/pixel. Each semantic change label has been annotated by a professional labeling group and classified into seven categories: non-change, non-vegetated ground surface, tree, low vegetation, water, buildings, and playgrounds. We adopt the standard split of 2,968 training pairs and 1,694 test pairs for our experiments following previous work.

The LandsatSCD is collected between the years 1990 and 2020 in Xinjiang, China. The dataset contains 8,468 pairs of bi-temporal image patches, most of which are generated through data augmentation. After filtering, a total of 2,385 image pairs are retained. Following previous work, we split the dataset into training, validation, and test sets with a ratio of 3:1:1, resulting in 1,431, 477, and 477 image pairs, respectively. Collected from Landsat satellite, the images in dataset are all in a resolution of 30m/pixel. All of them are annotated for changes across four land cover classes—farmland, desert, buildings, and water bodies.

### B. Implementation Details

All experiments are implemented using PyTorch and conducted on NVIDIA GeForce 4090. To ensure reproducibility, a fixed random seed is applied to all experiments. The proposed method is trained using stochastic gradient descent (SGD) optimizer with momentum. To ensure a robust feature representation, the ViT-G/16-1024 backbone is initialized with the official PerA weights, which were pre-trained on the large-scale RSRSD-5M dataset. The batch size is set to 8 and the weight decay is set to 1e-5. To improve training stability and efficiency, mixed-precision training is employed using automatic mixed precision (AMP) with gradient scaling. For SECOND dataset, the initial learning rate is set to 0.1 with a momentum of 0.9, and decays to zero following a polynomial schedule. A linear warm-up phase occupying the first 10% of total training iterations to enhance stability. For LandsatSCD dataset, due to its relatively simpler data distribution, we set the initial learning rate to 1.0 with a minimum learning rate of 0.5. Except for replacing SCLoss with our proposed SSCLoss, we utilize exactly the same loss functions as SCanNet to train our model. To mitigate overfitting during fine-tuning of large-scale models, we employ a DropPath rate of 0.3 along with data augmentation strategies including random flipping and color jittering. Due to the inherent differences in optimization objectives and training strategies across SCD methods, a fully unified training protocol is infeasible without altering the original design of each model. Therefore, we adopt the official implementations and recommended settings for all compared methods to preserve their intended behavior. Meanwhile, we ensure consistency in data splits, evaluation metrics, and preprocessing pipelines to maintain a fair comparison.

### C. Evaluation Metrics

Following previous SCD studies [36], [39], we evaluate the proposed method using four commonly adopted metrics: Overall Accuracy (OA), mean Intersection over Union (mIoU), Separated Kappa (SeK), and the SCD-targeted $F_1$ score ($F_{scd}$). Let $Q=\{q_{ij}\}$ denote the confusion matrix, where $q_{ij}$ is the number of pixels predicted as class $i$ while belonging to the ground-truth class $j$. The indices $i,j\in\{0,1,\ldots,N\}$, where 0 denotes the no-change class and $1,\ldots,N$ denote semantic change classes.

OA measures the ratio of correctly classified pixels to all pixels:

$$OA=\sum_{i=1}^{N} q_{ii} / \sum_{i=1}^{N}\sum_{j=0}^{N} q_{ij}. \quad (16)$$

Since unchanged pixels usually dominate SCD datasets, OA may not fully reflect the performance on changed regions. Therefore, we also report mIoU, which averages the IoU of the no-change class and the aggregated changed classes:

$$mIoU=(IoU_{nc}+IoU_c)/2 \quad (17)$$

where $IoU_{nc}$ and $IoU_c$ denote the average the IoU of the unchanged regions and changed regions:

$$IoU_{nc}=q_{00} / \sum_{i=0}^{N} q_{i0}+\sum_{j=0}^{N} q_{0j}-q_{00} \quad (18)$$

$$IoU_c=\sum_{i=1}^{N}\sum_{j=1}^{N} q_{ij} / \left(\sum_{i=0}^{N}\sum_{j=0}^{N} q_{ij}-q_{00}\right). \quad (19)$$

To further evaluate semantic discrimination in changed areas, we use SeK. SeK is computed based on a modified confusion matrix $\hat{Q}=\{\hat{q}_{ij}\}$, which is obtained from $Q$ by excluding the true no-change agreement term $q_{00}$, while preserving all other entries. In this way, the dominant correctly classified no-change pixels are not included in the agreement calculation, whereas false alarms, missed changes, and semantic confusions among changed classes are still considered:

$$\rho=\sum_{i=0}^{N} \hat{q}_{ii} / \left(\sum_{i=0}^{N}\sum_{j=0}^{N} \hat{q}_{ij}\right) \quad (20)$$

$$\eta=\sum_{i=0}^{N} \frac{\sum_{j=0}^{N} \hat{q}_{ij} * \sum_{j=0}^{N} \hat{q}_{ji}}{(\sum_{i=0}^{N}\sum_{j=0}^{N} \hat{q}_{ij})^2} \quad (21)$$

$$SeK=e^{IoU_c-1}\cdot(\rho-\eta)(1-\eta) \quad (22)$$

where $\rho$ is the observed agreement and η is the expected agreement by chance computed from $\hat{Q}$.

Finally, $F_{scd}$ is used to measure the semantic classification accuracy of changed pixels. It is calculated from the SCD precision $P_{scd}$ and recall $R_{scd}$:

$$P_{scd}=\sum_{i=1}^{N} q_{ii} / \sum_{i=1}^{N}\sum_{j=0}^{N} q_{ij} \quad (23)$$

$$R_{scd}=\sum_{i=1}^{N} q_{ii} / \sum_{i=0}^{N}\sum_{j=1}^{N} q_{ij} \quad (24)$$

$$F_{scd}=\frac{2*P_{scd}*R_{scd}}{P_{scd}+R_{scd}} \quad (25)$$

where $P_{scd}$ denotes the proportion of correctly predicted semantic change pixels among all pixels predicted as changed classes, and $R_{scd}$ denotes the proportion of correctly predicted semantic change pixels among all ground-truth changed pixels. These four metrics jointly evaluate overall accuracy, change localization, and semantic recognition performance in the SCD task.

*D. Comparison with other SCD methods*

We compare our proposed method with other SCD methods on standard benchmark datasets, including former state-of-the-art to demonstrate the effectiveness and generalization ability of our approach. These methods including: 1) classic SCD method HRSCD [21] which proposed basic paradigms; 2) SSCDl [36] and BiSRNet [36] which proposed SCLoss; 3) TED [27] and SCanNet [27] which integrated transformer; and 4) Mamba-FCS [30] which is recently published SCD method that utilizes the Mamba architecture. For fair comparison, all methods adopted experimental setup in the original paper and are evaluated under the same preprocessing and training settings whenever possible. We report best results on the test sets and all experiments are conducted using the same training/test split and input resolution.

As shown in Table I, our method outperforms the compared approaches across all metrics. On the challenging SECOND dataset, PerASCD improves the $F_{scd}$ by 2.88% and the SeK by 2.93% over the baseline SCanNet. Similarly, on the LandsatSCD dataset, it achieves significant gains of 4.01% in $F_{scd}$ and 10.44% in SeK compared to the baseline. When compared to the previous state-of-the-art Mamba-FCS, our method maintains a consistent lead, improving the $F_{scd}$ by 0.63% and 1.63% on the SECOND and LandsatSCD datasets, respectively. Notably, the superiority in the SeK metric is even more pronounced, with improvements of 0.61% (SECOND) and 4.95% (LandsatSCD) over Mamba-FCS. These performance gains can be primarily attributed to three core factors: first, the robust semantic representations provided by the PerA foundation model; second, the sophisticated feature arbitration and multi-scale fusion facilitated by the CG-Decoder, particularly through the dynamic gating mechanism of the CAGM; and third, the enhanced optimization stability ensured by the SSCLoss. Specifically, the substantial improvement in SeK highlights the CG-Decoder's superior capability in accurately resolving complex semantic transitions, which benefits from the synergy between high-level foundation model features and our adaptive decoding interface.

Fig. 6(a) and Fig. 6(b) illustrate qualitative comparisons on SECOND dataset and LandsatSCD dataset respectively. To better highlight the superiority of our method, we selected test examples from the two datasets that exhibit diverse land cover types, complex objects, and demand fine-grained predictions. Our approach produces more precise boundaries and reduces false positives in complex change regions compared to all other methods. In scenarios that require fine-grained prediction, our model generally outperforms all other methods, clearly separating multiple buildings or delineating the contours of changing rivers. In the third example from the SECOND dataset, our model is the only one that successfully identifies the otherwise inconspicuous tennis court next to the playground, demonstrating the advantages of semantic understanding provided by RS foundation model.

To provide a more fine-grained qualitative analysis, Fig. 7 presents representative comparisons between SCanNet and PerASCD on the SECOND and LandsatSCD datasets. In the first group of Fig. 7(a), the red boxes show that our method produces clearer separation between adjacent building instances and preserves more complete object structures than the baseline. The cyan and magenta boxes highlight regions affected by strong vegetation phenology and radiometric variations. Although these appearance differences can easily be confused with real semantic changes, PerASCD maintains more accurate semantic predictions and suppresses most pseudo-changes. This improvement benefits from the temporally consistent semantic representations learned by the PerA backbone and the adaptive pseudo-change suppression provided by the CG-Decoder. In the second group, the red box shows that our model generates smoother and more regular object boundaries. The yellow and magenta boxes indicate that the model correctly recognizes low vegetation under different surface backgrounds and spectral conditions, while the blue box shows that our method successfully distinguishes a playground from spectrally similar low vegetation. These results demonstrate the advantages of global semantic understanding and coarse-to-fine multi-scale feature fusion in resolving confusing land-cover categories. Fig. 7(b) further provides a local zoom-in comparison on the medium-resolution LandsatSCD dataset. Compared with baseline SCanNet, PerASCD produces more compact and regular boundaries, and achieves closer spatial agreement with the ground truth. These examples confirm that our method improves boundary delineation, pseudo-change suppression, and semantic consistency across different spatial resolutions and scene conditions.

The comparative experiments demonstrate the superior performance of our proposed method, showing clear quantitative improvements on both simple and complex datasets, as well as qualitative advantages in boundary delineation, pseudo-change suppression, and semantic consistency, which validates the effectiveness of our approach.

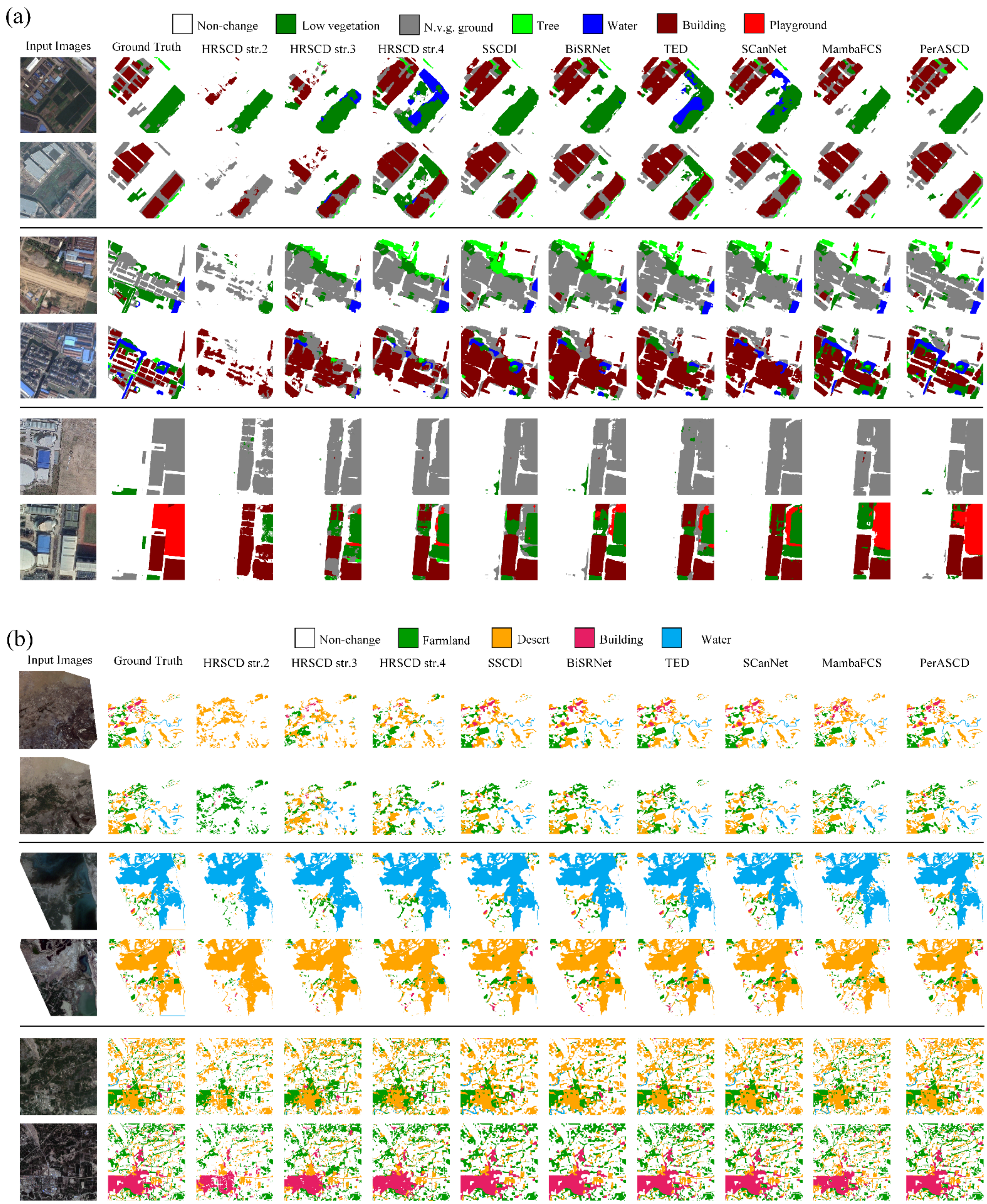

**Fig. 6.** Examples of results provided by different methods in the comparative experiments. (a) Results on the SECOND dataset; (b) Results on the LandsatSCD dataset.

TABLE I
COMPARISON WITH OTHER SCD METHODS, * INDICATES RESULTS REPRODUCED FROM PAPERS AND BOLD INDICATES THE BEST RESULT.

| Method | SECOND | | | | LandsatSCD | | | |
|---|---|---|---|---|---|---|---|---|
| | OA | $F_{scd}$ | mIoU | SeK | OA | $F_{scd}$ | mIoU | SeK |
| HRSCD Str. 2 | 85.69 | 51.24 | 65.05 | 11.60 | 90.67 | 70.71 | 73.55 | 25.09 |
| HRSCD Str. 3 | 85.33 | 54.05 | 66.91 | 13.37 | 91.17 | 72.55 | 77.22 | 29.64 |
| HRSCD Str. 4 | 87.25 | 60.22 | 71.97 | 20.13 | 91.96 | 74.77 | 79.32 | 33.39 |
| SSCDl | 87.25 | 62.72 | 73.02 | 22.75 | 94.32 | 83.59 | 84.31 | 47.60 |
| BiSRNet | 87.67 | 63.10 | 73.29 | 23.05 | 94.13 | 82.97 | 83.75 | 46.23 |
| TED | 87.49 | 62.94 | 73.06 | 22.67 | 95.43 | 86.62 | 87.12 | 54.69 |
| SCanNet | 87.57 | 63.53 | 73.03 | 23.18 | 95.42 | 86.89 | 86.96 | 54.77 |
| Mamba-FCS* | 88.62 | 65.78 | 74.07 | 25.50 | 96.25 | 89.27 | 88.81 | 60.26 |
| PerASCD | **88.70** | **66.41** | **74.33** | **26.11** | **96.84** | **90.90** | **90.54** | **65.21** |

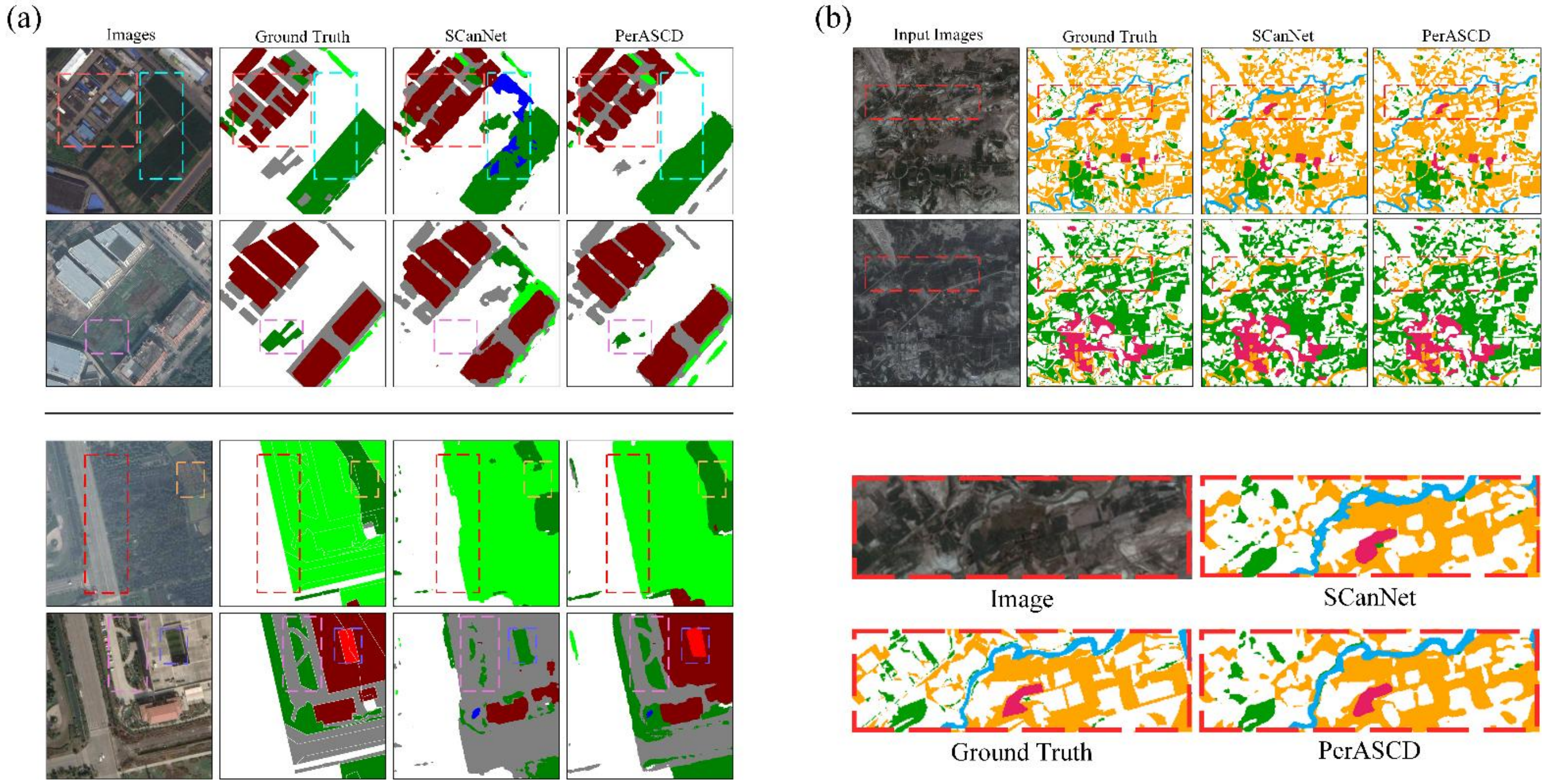

**Fig. 7.** Fine-grained visual comparison and local zoom-in analysis on the SECOND and LandsatSCD Datasets: (a) Comparison on representative samples from the SECOND dataset. (b) Local zoom-in comparison on the LandsatSCD dataset.

*E. Ablation studies*

We conduct a series of ablation studies to systematically evaluate the contributions and behavior of the proposed PerASCD framework, including the adaptability of the CG-Decoder to different backbone encoders, the effectiveness of individual components, data efficiency under varying training-set sizes, numerical stability, and the influence of the temperature parameter $\tau$. To ensure fair and controlled evaluation, each experiment follows the same base implementation and training protocol unless otherwise specified, while only the factor under investigation is varied.

Specifically, all configurations trained in our ablation experiments are implemented in PyTorch and trained on NVIDIA GeForce RTX 4090 GPUs using the same fixed random seed. They are optimized using Stochastic Gradient Descent (SGD) with a momentum of 0.9 and a weight decay of 1e-5. We use the same batch size of 8, mixed-precision training (AMP), a 10% linear warm-up phase, and an initial learning rate of 0.1 with cosine decay. The same loss formulation and data augmentation strategy are adopted unless the corresponding ablation explicitly investigates these factors. For experiments on LandsatSCD, the dataset-specific optimization settings described in Section IV-B are retained. Thus, throughout the ablation studies, all non-target variables are kept consistent whenever applicable, allowing the contribution of each investigated factor to be assessed under a controlled setting.

1) *Ablation Study with Different Backbone Encoders:* To evaluate the cross-backbone adaptability of the proposed CG-Decoder under a fair and controlled setting, we explicitly separate backbone-dependent factors from downstream SCD training factors. Across the CG-Decoder-based configurations in Table II, the downstream framework and training protocol are kept consistent, including the decoder architecture, optimization objective, optimizer, learning-rate schedule, batch size, warm-up strategy, mixed-precision setting, data augmentation strategy, random seed, and hardware environment. The same CG-Decoder configuration, including the number of decoding blocks and internal feature dimension, is retained across different backbones; only the input projection layers are adjusted when necessary to accommodate backbone-specific feature dimensions.

Each encoder is initialized using its officially released pre-trained weights and corresponding normalization statistics. Following the controlled-evaluation principle discussed in [41], we retain only those backbone-specific settings that are intrinsic to the official implementations rather than imposing identical settings that may be inappropriate for heterogeneous architectures. Accordingly, the DropPath rate is generally set to 0.3, while being adjusted to 0.1 for SeCo and 0 for SatMAE and Scale-MAE to preserve their intended operating conditions. These adjustments do not modify the downstream SCD architecture or optimization objective.

Under this controlled setting, the encoder and its associated official pre-trained representation constitute the primary experimental variables. We integrate the CG-Decoder with representative CNN-, Transformer-, and state-space-model backbones, including ResNet [42], SeCo [9], Swin [43], ViT-based SatMAE [44], Scale-MAE [8], and VMamba [45].

Therefore, the CG-Decoder-based configurations in Table II constitute a controlled backbone comparison, where the encoder and its associated official pre-trained representation are treated as the primary experimental variables, while the downstream SCD framework and training protocol remain consistent. This design provides a fair basis for evaluating whether the proposed CG-Decoder can maintain its effectiveness across heterogeneous backbone architectures and pre-training paradigms, rather than benefiting from a particular downstream configuration.

As shown in Table II, the proposed framework demonstrates seamless adaptation and strong generalization across this wide range of architectures. Notably, even when equipped with a classic, lightweight CNN backbone like ResNet50, the CG-Decoder achieves highly competitive results, underscoring its broad computational applicability for resource-constrained scenarios. It is worth highlighting that, with the sole exception of the SeCo-pre-trained variant, all backbones integrated with our CG-Decoder consistently outperform the baseline SCanNet across all major metrics. Meanwhile, the successful integration of SatMAE and Scale-MAE further demonstrates that the proposed decoder can effectively accommodate remote sensing backbones pre-trained using masked image modeling. High-capacity models like Swin-L exhibit robust representation capabilities, further highlighting the structural flexibility of our decoder.

A particularly crucial observation arises from the integration of VMamba-Base, the exact backbone utilized in the recently published Mamba-FCS [30]. While Mamba-FCS relies on a highly specialized architecture meticulously tailored specifically for Mamba-based feature extraction, our CG-Decoder, without any backbone-specific structural optimization, achieves performance highly comparable to this recent state-of-the-art method. This finding fundamentally addresses the unique advantage of our approach: the core contribution of the CG-Decoder lies not in marginal numerical superiority over specialized networks, but in its profound architectural adaptability. It proves that our unified gating mechanism can efficiently unlock the potential of emerging architectures out-of-the-box, matching the efficacy of custom-designed networks while maintaining a unified, backbone-agnostic pipeline.

TABLE II
ABLATION STUDY WITH DIFFERENT BACKBONE ENCODERS, * INDICATES RESULTS REPRODUCED FROM PAPERS AND BOLD INDICATES THE BEST RESULT.

| Encoder/Method | | SECOND | | | |
|---|---|---|---|---|---|
| | | OA | $F_{scd}$ | mIoU | SeK |
| SCanNet | | 87.82 | 64.30 | 73.31 | 23.84 |
| Mamba-FCS* | | 88.62 | 65.78 | 74.07 | 25.50 |
| ResNet50 | w/ CG-Decoder | 87.77 | 64.81 | 73.63 | 24.63 |
| VMamba-B | w/ CG-Decoder | 88.37 | 65.61 | 74.01 | 25.31 |
| Swin-L | w/ CG-Decoder | 88.19 | 65.74 | 73.91 | 25.40 |
| ResNet50 (SeCo pre-trained) | w/ CG-Decoder | 87.58 | 63.21 | 72.95 | 22.79 |
| ViT-L (SatMAE pre-trained) | w/ CG-Decoder | 88.15 | 64.97 | 73.59 | 24.51 |
| ViT-L (Scale-MAE pre-trained) | w/ CG-Decoder | 88.30 | 64.90 | 73.69 | 24.56 |
| ViT-G/16/1024 (PerA pre-trained) | w/ CG-Decoder | **88.70** | **66.41** | **74.33** | **26.11** |

Regarding pre-training strategies, RS pre-training does not consistently outperform ImageNet pre-training across all settings. In particular, the SeCo pre-trained ResNet50 exhibits slightly inferior performance compared to its ImageNet pre-trained counterpart. This can be explained by two factors: first, SeCo is pre-trained on Sentinel-2 imagery, whose relatively lower spatial resolution may limit its suitability for high-resolution SCD; second, the relatively modest scale of the SeCo pre-training dataset may constrain the richness of the learned representations. In contrast, the model pre-trained with PerA achieves the best overall performance. Benefiting from a large-scale RS pre-training dataset and a powerful foundation model architecture, PerA provides highly transferable semantic representations that perfectly synergize with the CG-Decoder. These results suggest that the combination of large-scale RS foundation models and an adaptive, plug-and-play decoding interface is the optimal trajectory for fully unlocking performance gains in SCD.

TABLE III
COMPUTATIONAL COMPLEXITY UNDER DIFFERENT BACKBONE-DECODER CONFIGURATIONS.

| Method | Params. (M) | FLOPs (G) | Mem. (GB) | Infer. (ms/pair) | Train (ms/iter) |
|---|---|---|---|---|---|
| ResNet50+ SCanNet Decoder | 31.51 | 263.54 | 2.44 | 15.25 ±0.12 | 66.73 ±2.04 |
| ResNet50+ CG-Decoder | 32.44 | 159.34 | 1.61 | 18.32 ±0.04 | 69.17 ±1.16 |
| PerA+ SCanNet Decoder | 547.18 | 2898.19 | 15.66 | 200.60 ±1.64 | 540.62 ±14.81 |
| PerA+CG-Decoder | 548.17 | 2853.05 | 15.09 | 182.67 ±13.87 | 530.58 ±37.84 |

We further conducted a computational complexity comparison under different backbone–decoder configurations, as reported in Table III. All runtime results are averaged over five independent runs and presented as mean ± standard deviation. Under the ResNet50 backbone, replacing the SCanNet decoder with the CG-Decoder only slightly increases the parameter count from 31.51 M to 32.44 M, while substantially reducing FLOPs from 263.54 G to 159.34 G and peak training memory from 2.44 GB to 1.61 GB. Although the inference and training times increase moderately to 18.32 ms/pair and 69.17 ms/iter, respectively, the notable reductions in computational operations and memory consumption demonstrate the efficient multi-scale fusion capability of the CG-Decoder. Under the higher-capacity PerA backbone, the CG-Decoder similarly introduces only a marginal parameter increase from 547.18 M to 548.17 M, while reducing FLOPs, memory consumption, inference time, and training time from 2898.19 G, 15.66 GB, 200.60 ms/pair, and 540.62 ms/iter to 2853.05 G, 15.09 GB, 182.67 ms/pair, and 530.58 ms/iter, respectively. These results show that the CG-Decoder achieves favorable computational and memory efficiency without substantially increasing the model size. Its consistent advantages across both lightweight and large-scale backbones further demonstrate that the proposed adaptive multi-scale fusion design can improve feature decoding efficiency while maintaining broad architectural applicability.

2) *Ablation Study with Different Individual Improvements:* To investigate the contribution of each individual improvement in our method, we conduct an ablation study by selectively enabling or disabling each component. As shown in Table IV, we conducted ablation experiments on the pre-trained encoder, CG-Decoder, CAGM, and SSCLoss individually. All experiments were performed under the same settings, and the results are reported in terms of OA, $F_{scd}$, mIoU, and SeK on the SECOND dataset.

From the experimental results, we can observe that performance improvements are not limited to using the PerA pre-trained ViT. When combined with the CG-Decoder, a ResNet-50 backbone also achieves consistent gains across all metrics compared to the baseline SCanNet. Furthermore, the contribution of CAGM is highly pronounced: incorporating CAGM leads to a notable improvement of 0.79% in the $F_{scd}$. This performance boost indicates that CAGM functions as an effective dynamic feature arbiter, implicitly learning to prioritize change-relevant information while suppressing environmental noise and pseudo-changes across different scales. By adaptively gating the feature flow, CAGM enhances the model's discriminative power in complex scenarios without relying on rigid, handcrafted supervision, thereby bridging the gap between high-level semantic depth and fine-grained change localization. In addition, compared with the original SCLoss, SSCLoss does not bring substantial performance enhancement. Instead, its primary benefit lies in enhancing the training stability and robustness of the model, as we discussed in Section III-D and further validated in item 4) of this section.

To provide insights into the interpretability of CAGM, we visualize its learned gating weight maps. As shown in Fig. 8, for each group of examples, the first and second rows present the original images, ground-truth, and model predictions, while the third and fourth rows illustrate the weight heatmaps of the deep features (low resolution) and shallow features (high resolution) learned by the CAGM in every CG-Decoder Block.

From deep to shallow layers, we observe that the heatmaps align well with our design expectations. Although no explicit supervision is imposed to enforce the learning of correct change regions, the weight heatmaps predicted by the CAGM in each CG-Decoder Block generally delineate the boundary between changed and unchanged areas. As the feature maps are progressively upsampled in spatial resolution, the heatmaps become increasingly refined, with more accurate focus regions. We also experimented with using the ground-truth change masks as explicit learning targets; however, such rigid constraints instead limited the model's performance.

TABLE IV
ABLATION STUDY WITH DIFFERENT INDIVIDUAL IMPROVEMENTS, BOLD INDICATES THE BEST RESULT.

| Method | SECOND | | | |
|---|---|---|---|---|
| | OA | $F_{scd}$ | mIoU | SeK |
| SCanNet | 87.82 | 64.30 | 73.31 | 23.84 |
| ResNet50+CG-Decoder | 87.77 | 64.81 | 73.63 | 24.63 |
| PerASCD (SCanNet Decoder) | 88.35 | 65.95 | 74.14 | 25.72 |
| PerASCD w/o CAGM | 88.48 | 65.62 | 74.16 | 25.46 |
| PerASCD w/ $L_{SC}$ | **88.87** | 66.39 | **74.36** | 26.07 |
| PerASCD | 88.70 | **66.41** | 74.33 | **26.11** |

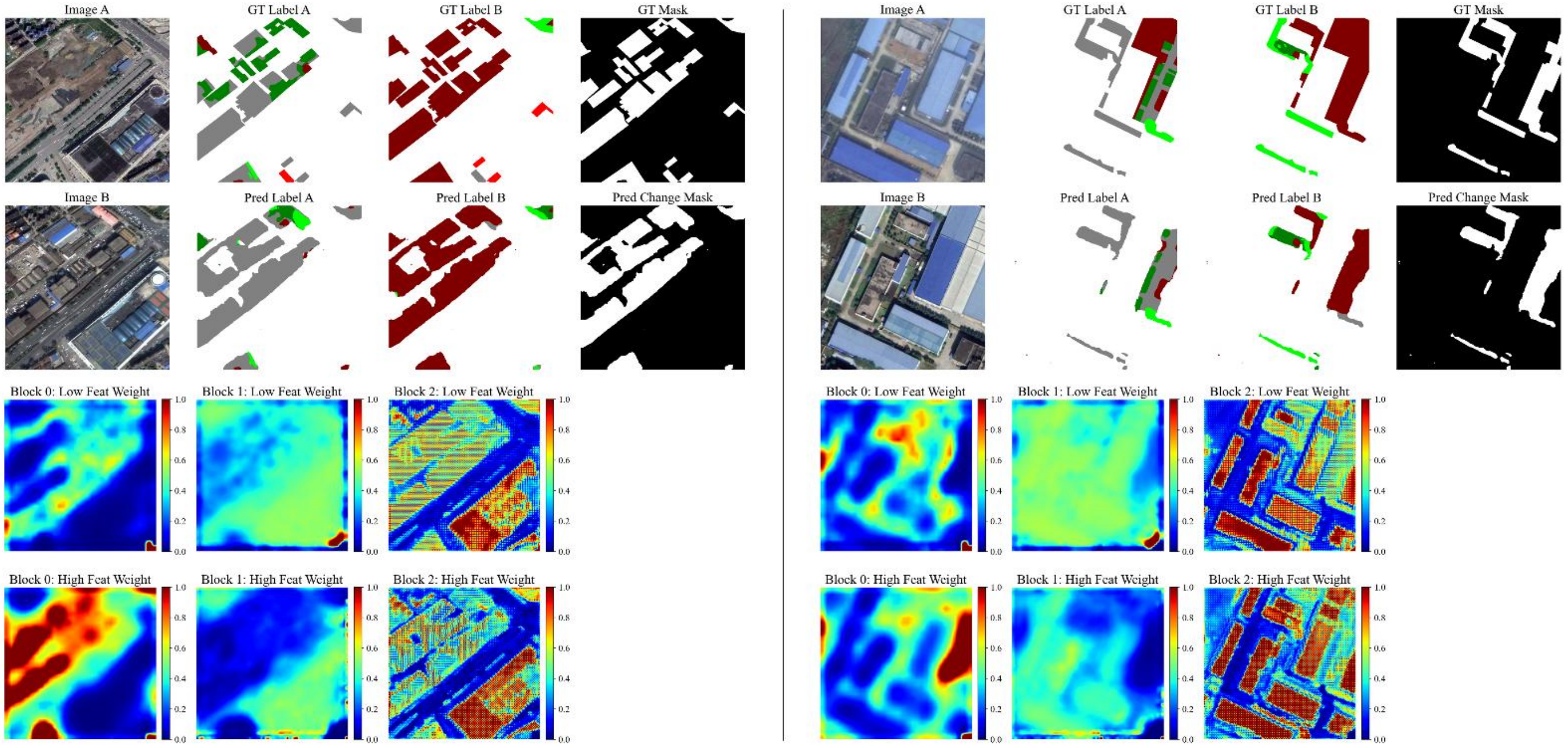

**Fig. 8.** Examples comparing CAGM weight heatmaps with the corresponding prediction results.

From a semantic perspective, the learning process exhibits a clear progression from coarse to fine. In most cases, the model first roughly localizes the change regions using highly compressed deep semantic features, then learns the relatively simpler unchanged regions, and finally refines its attention to the change regions to capture fine-grained details. Notably, in the heatmaps of the final layers, the genuine change regions are almost precisely outlined, yet they typically do not correspond to the highest-attention (red) areas. Instead, they are concentrated in grid-like striped regions with moderate weight values, ranging from 0.3 to 0.7. We hypothesize that this counter-intuitive phenomenon fundamentally reflects the module's mechanism for suppressing pseudo-changes. Because the learned gating weights are jointly applied to regulate both semantic and change representations, the CAGM actively allocates its highest attention (the red regions) to confusing non-change objects that are highly susceptible to being misclassified. These areas typically exhibit significant radiometric differences caused by environmental variations—such as illumination discrepancies, seasonal shifts, or sensor differences—without any actual semantic category transition. By assigning extreme weights to these pseudo-changes, the CAGM enforces strong semantic consistency to actively filter out false alarms, proving its role as an intelligent feature arbiter rather than a simple change highlight map.

3) *Ablation Study on Decoder Design under Varying Training Data Scales*: To further investigate the robustness of our method and validate the capability of the foundation model in reducing reliance on large-scale annotated training samples, we conduct an ablation study under varying training data scales on the SECOND dataset. We randomly sample 10%, 50%, and 100% of the training data to train the models, and the results are reported in Table V.

From the experimental results, three significant observations can be drawn. First, the proposed CG-Decoder consistently demonstrates robustness against severe class imbalance. In SCD datasets, the background (unchanged) class is overwhelmingly large, which often biases the predictions. Compared to the baseline decoder (SCanNet Decoder), our CG-Decoder effectively mitigates this issue, enabling much more accurate predictions for small change categories. Consequently, the three most critical metrics for SCD ($F_{scd}$, mIoU, and SeK) exhibit significant improvements across all data scales. Second, the well pre-trained RS foundation model plays an indispensable role in preventing pseudo-changes. When the PerA foundation model is removed (i.e., the standard SCanNet), the performance drops drastically at all data scales. This highlights that the RS foundation model provides superior semantic understanding and spatio-temporal invariance, which are crucial for filtering out irrelevant environmental noises and pseudo-changes, especially when the available training data is extremely limited (e.g., at the 10% scale).

Finally and most notably, the foundation model introduces favorable transferability and data efficiency, effectively breaking the bottleneck of high annotation costs. As clearly shown in Table V, our PerASCD (PerA + CG-Decoder) trained with only 50% of the data achieves an $F_{scd}$ of 64.79% and a SeK of 24.21%. Remarkably, this performance still outperforms the baseline SCanNet trained with 100% of the data ($F_{scd}$ of 64.30%, SeK of 23.84%). Even with only 10% of the training data, PerASCD maintains a highly competitive performance. This compelling evidence perfectly demonstrates that utilizing an RS foundation model can effectively break the bottleneck of high annotation costs in practical SCD applications.

TABLE V
ABLATION STUDY ON DECODER DESIGN UNDER VARYING TRAINING DATA SCALES, BOLD INDICATES THE BEST RESULT IN EACH EXPERIMENTAL GROUP.

| Method | Training ratio | SECOND | | | |
|---|---|---|---|---|---|
| | | OA | $F_{scd}$ | mIoU | SeK |
| PerA+CG-Decoder | 100% | **88.70** | **66.41** | **74.33** | **26.11** |
| PerA+SCanNet Decoder | | 88.35 | 65.95 | 74.14 | 25.72 |
| SCanNet | | 87.82 | 64.30 | 73.31 | 23.84 |
| PerA+CG-Decoder | 50% | 87.65 | **64.79** | **73.11** | **24.21** |
| PerA+SCanNet Decoder | | **88.04** | 64.40 | 72.99 | 23.61 |
| SCanNet | | 87.12 | 62.30 | 72.04 | 21.63 |
| PerA+CG-Decoder | 10% | 86.32 | **60.25** | **70.64** | **19.29** |
| PerA+SCanNet Decoder | | **86.71** | 59.91 | 70.56 | 18.94 |
| SCanNet | | 84.86 | 56.95 | 68.25 | 15.72 |

4) *Ablation Study of Numerical Instability on LandsatSCD Dataset:* We conduct detailed ablation experiments to explore the inner mechanism of the numerical instability and validate the effect of our proposed SSCLoss. Due to the relatively simple data distribution of LandsatSCD and the sufficient capacity and structural stability of the PerASCD, the network is able to tolerate a relatively large learning rate. Empirically, we observe that PerASCD achieves its optimal performance on LandsatSCD with a learning rate as high as 1.0. However, we also observe that models incorporating the original SCLoss may suffer from sporadic numerical instability. Moreover, the probability of such instability increases as the learning rate becomes larger due to the larger update step size. With an initial learning rate of 1.0 and a minimum of 0.5, the model achieves peak accuracy. However, numerical instability is nearly unavoidable, even under strict gradient clipping. In such cases, isolated unstable batches can trigger sudden, large drops in the loss and gradient values, manifesting as cliff-like deviations in the training curves. Additionally, we observe that using FP32 significantly reduces numerical instability. This is because while mixed-precision training lowers computational cost, it also increases the likelihood of gradient underflow, making small differences near the hinge margin more prone to being rounded to zero, which in turn sharply raises the probability of instability in the subsequent iterations.

During the early stages of training, we also observe fluctuations in the training curves that resemble numerical instability, but do not exhibit the characteristic cliff-like behavior. These fluctuations are not caused by gradient discontinuities near the margin, but rather stem from the randomness introduced by weight initialization. To ensure a fair and rigorous evaluation, we ignore such fluctuations occurring within the first five epochs. Numerical instability is defined as the occurrence of a cliff-like dip in performance during the subsequent training process. In multiple runs, the presence of numerical instability is determined if such an event occurs at least once.

As shown in the Table VI, we can clearly observe that the numerical instability issues are strongly correlated with floating-point precision and loss function. On the LandsatSCD dataset, when the maximum learning rate is set to 1.0 and SCLoss is adopted together with FP16 training, numerical instability is frequently observed, regardless of whether gradient clipping is applied. After disabling SCLoss, although the overall performance drops significantly, such instability is no longer observed. When we replace the loss function with our proposed SSCLoss ($\tau = 0.5$), the issue is mitigated significantly with gradient clipping. However, when gradient clipping is disabled, numerical instability re-emerges. This observation indicates that although the proposed SSCLoss is effective in mitigating gradient instability, it is insufficient on its own to fully stabilize gradient fluctuations under a high learning rate. Additionally, when FP32 training is adopted, numerical instability is rarely observed, which is consistent with our hypothesis that small gradients below the margin are more likely to underflow under mixed-precision training.

TABLE VI
ABLATION STUDY OF NUMERICAL INSTABILITY ON LANDSATSCD DATASET. GC DENOTES GRADIENT CLIPPING AND NI DENOTES NUMERICAL INSTABILITY.

| Method | GC | Precision | Loss | $F_{scd}$ | NI |
|---|---|---|---|---|---|
| PerASCD | 1.5 | FP16 | - | 80.39 | No |
| PerASCD | 1.5 | FP16 | $L_{SC}$ | - | Yes |
| PerASCD | - | FP16 | $L_{SC}$ | - | Yes |
| PerASCD | - | FP32 | $L_{SC}$ | 90.48 | No |
| PerASCD | 1.5 | FP16 | $L_{SSC}$ | 90.90 | No |
| PerASCD | - | FP16 | $L_{SSC}$ | - | Yes |
| PerASCD | - | FP32 | $L_{SSC}$ | 90.52 | No |
| PerASCD | - | FP16 | - | 79.19 | No |

5) *Ablation Study of the Temperature Parameter $\tau$:* We also conduct a bunch of ablation experiments to explore the effect of temperature parameter $\tau$ on both LandsatSCD and SECOND. The results indicate that the temperature parameter $\tau$ should be adjust according the distribution of the training dataset.

As shown in Table VII, since the LandsatSCD dataset exhibits a simpler data distribution with less noise, the model is less prone to overfitting. As a result, the best performance is achieved at a relatively high temperature value ($\tau = 0.5$), as weaker constraints allow the model to more effectively capture fine-grained details in a low-noise data regime. When temperature value $\tau$ achieved 0.01, the numerical instability is observed due to a strong constraint with high upgrade step size.

TABLE VII
ABLATION STUDY OF THE TEMPERATURE PARAMETER TAU ON LANDSATSCD DATASET

| Method | $\tau$ | OA | $F_{scd}$ | mIoU | SeK |
|---|---|---|---|---|---|
| PerASCD | 0.01 | 96.48 | 89.85 | 89.60 | 62.31 |
| PerASCD | 0.05 | 96.67 | 90.39 | 90.06 | 63.73 |
| PerASCD | 0.1 | 96.79 | 90.72 | 90.38 | 64.69 |
| PerASCD | **0.5** | **96.84** | **90.90** | **90.54** | **65.21** |
| PerASCD | 1 | 96.75 | 90.60 | 90.26 | 64.33 |

On the SECOND dataset, as shown in Table VIII, due to the complexity of SECOND dataset, a high temperature involved more noise information, leading to a limited performance. As the temperature $\tau$ is gradually decreased, the optimal $F_{scd}$ performance consistently improves and reaches its peak at $\tau = 0.01$ or $\tau = 0.02$. Since a relatively smaller initial learning rate of 0.1 is adopted during training, the strong regularization imposed by SSCLoss, although closely resembling the original SCLoss, does not lead to frequent numerical instability. In contrast, benefiting from improved regularization and smoother gradients, overfitting is substantially reduced, leading to a significant improvement in overall performance.

TABLE VIII
ABLATION STUDY OF THE TEMPERATURE PARAMETER TAU ON SECOND DATASET, BOLD INDICATES THE BEST RESULT.

| Method | $\tau$ | OA | $F_{scd}$ | mIoU | SeK |
|---|---|---|---|---|---|
| PerASCD | 0.005 | 88.30 | 66.33 | 74.30 | 26.17 |
| **PerASCD** | **0.01** | **88.70** | 66.41 | 74.33 | 26.11 |
| **PerASCD** | **0.02** | 88.65 | **66.45** | **74.45** | **26.23** |
| PerASCD | 0.1 | 88.34 | 66.36 | 74.32 | 26.21 |
| PerASCD | 0.5 | 88.53 | 66.33 | 74.33 | 26.10 |
| PerASCD | 1 | 88.36 | 66.32 | 74.28 | 26.12 |

## V. CONCLUSION AND DISCUSSION

### *A. Conclusion*

In this paper, we introduced PerASCD, a robust and unified SCD framework that synergizes the representational power of RS foundation models with a streamlined, modular Cascaded Gated Decoder (CG-Decoder). Our research demonstrates that the bottleneck of semantic change detection lies not only in feature extraction but in the intelligent arbitration of multi-scale semantic and change information. To this end, we propose a modular CG-Decoder that processes multi-scale features in a coarse-to-fine manner and seamlessly adapts to diverse vision encoders, in which a Change-Aware Gating Module (CAGM) intelligently arbitrates change information to effectively localize and suppress pseudo-changes, along with a numerically stable supervision objective, jointly improving the model's accuracy, data efficiency, cross-backbone adaptability, and interpretability. Experiments on the SECOND and LandsatSCD benchmarks support the following findings under the evaluated settings:

1) Improved **Performance and Data Efficiency:** Quantitatively, PerASCD achieved a SeK of 26.11% on the SECOND dataset and 65.21% on LandsatSCD, surpassing the previous state-of-the-art Mamba-FCS by 0.61% and 4.95%, respectively. Furthermore, the model demonstrated favorable data efficiency: when trained with only 50% of the available data, PerASCD (64.79% $F_{scd}$ and 24.21% SeK) still outperformed the baseline SCanNet trained on the full 100% dataset (64.30% $F_{scd}$ and 23.84% SeK). These results indicate the potential of the proposed framework to reduce its reliance on labeled training samples under the evaluated setting.
2) **Cross-Backbone Adaptability:** The CG-Decoder demonstrates broad architectural adaptability across the evaluated backbone families. By integrating it with diverse backbones, from the classic ResNet-50 to the emerging VMamba-B, the proposed decoder achieves competitive performance without requiring backbone-specific structural redesign. In particular, when combined with VMamba-B, the same backbone architecture adopted by Mamba-FCS, the CG-Decoder achieves performance highly comparable to Mamba-FCS. These results indicate that the proposed decoder can effectively accommodate heterogeneous backbone representations under the evaluated configurations.
3) **Enhanced Interpretability:** Visualization of the learned CAGM gating maps provides insight into how the model regulates semantic and change features across different decoding stages. The observed gating patterns are generally consistent with the intended design, showing increased responses in regions susceptible to pseudo-changes and progressive refinement of change-related structures across scales. These observations suggest that the CAGM contributes to suppressing confusing radiometric and environmental variations while maintaining semantic consistency.
4) **Improved Numerical Stability:** The SSCLoss is introduced to address the numerical instability of SCLoss under mixed-precision training in the SCD task. The experimental results show that, when combined with gradient clipping, SSCLoss provides a more stable optimization process under the tested high-learning-rate configuration on LandsatSCD, while maintaining competitive prediction performance. These results support the effectiveness of SSCLoss as a more stable semantic consistency constraint under the evaluated training conditions.

### *B. Discussion*

Beyond the demonstrated advantages, several aspects of PerASCD merit further discussion and open up directions for future work:

1) **On Dataset Coverage and Generalization:** Although SECOND and LandsatSCD provide two complementary SCD scenarios with substantial differences in spatial resolution, scene characteristics, and imaging conditions, the current evaluation is still limited to these two public benchmarks. Therefore, the generalization conclusions of this study should be interpreted within the scope of the evaluated datasets and backbone configurations. Further validation on additional SCD datasets covering different geographic regions, sensors, spatial resolutions, seasonal conditions, and land-cover distributions will be necessary to more comprehensively assess the robustness and transferability of the proposed framework.

2) **On Multimodal and LLM Adaptability:** Owing to its modular design, the CG-Decoder can be conceptually regarded as a flexible change-detection head, analogous to a classifier or segmentation head. The cross-backbone results obtained in this study suggest that such a design may potentially be extended to a broader range of vision encoders, including architectures designed for multimodal data or LLM-based representations. However, its compatibility with substantially different modalities, such as SAR–optical fusion, or with LLM-driven semantic representations has not yet been experimentally validated. Therefore, systematically evaluating the adaptability of the proposed decoder in these settings constitutes an important direction for future work.
3) **On the Hyper parameter τ in SSCLoss:** SSCLoss mitigates the numerical instability observed with SCLoss under the evaluated mixed-precision and high-learning-rate settings, but introduces an additional hyperparameter $\tau$ that controls the temperature scaling. In practice, $\tau$ requires dataset-dependent tuning for optimal performance, which may hinder straightforward application across diverse settings. Developing an adaptive scheduling strategy for $\tau$, or designing a hyperparameter-free variant that retains the stabilization effect, remains a relevant direction for future work.
4) **On Deployment and Model Compression:** PerASCD achieves state-of-the-art results when powered by large pre-trained models, yet this comes with a substantial computational footprint that limits deployment on resource-constrained edge devices. Lightweight encoders already yield respectable performance, but bridging the remaining gap without the full model could be pursued via knowledge distillation, pruning, or structural re-parameterization. These techniques offer promising routes to a highly efficient yet accurate SCD solution.

## References

[1] S. Tian, A. Ma, Z. Zheng, and Y. Zhong, “Hi-UCD: A Large-scale Dataset for Urban Semantic Change Detection in Remote Sensing Imagery,” Dec. 28, 2020, *arXiv*: arXiv:2011.03247. doi: 10.48550/arXiv.2011.03247.

[2] A. Ochtyra, A. Marcinkowska-Ochtyra, and E. Raczko, “Threshold- and trend-based vegetation change monitoring algorithm based on the inter-annual multi-temporal normalized difference moisture index series: A case study of the Tatra Mountains,” *Remote Sensing of Environment*, vol. 249, p. 112026, Nov. 2020, doi: 10.1016/j.rse.2020.112026.

[3] Z. Zheng, Y. Zhong, J. Wang, A. Ma, and L. Zhang, “Building damage assessment for rapid disaster response with a deep object-based semantic change detection framework: From natural disasters to man-made disasters,” *Remote Sensing of Environment*, vol. 265, p. 112636, Nov. 2021, doi: 10.1016/j.rse.2021.112636.

[4] K. Cha, J. Seo, and T. Lee, “A Billion-scale Foundation Model for Remote Sensing Images,” *IEEE J. Sel. Top. Appl. Earth Observations Remote Sensing*, pp. 1–17, 2024, doi: 10.1109/JSTARS.2024.3401772.

[5] X. Sun *et al.*, “RingMo: A Remote Sensing Foundation Model With Masked Image Modeling,” *IEEE Trans. Geosci. Remote Sensing*, vol. 61, pp. 1–22, 2023, doi: 10.1109/TGRS.2022.3194732.

[6] U. Mall, B. Hariharan, and K. Bala, “Change-Aware Sampling and Contrastive Learning for Satellite Images,” in *2023 IEEE/CVF Conference on Computer Vision and Pattern Recognition (CVPR)*, Vancouver, BC, Canada: IEEE, Jun. 2023, pp. 5261–5270. doi: 10.1109/CVPR52729.2023.00509.

[7] H. Shen, H. Gu, H. Li, Y. Yang, and A. Qiu, “A contrastive learning foundation model based on perfectly aligned sample pairs for remote sensing images,” *Geo-spatial Information Science*, vol. 0, no. 0, pp. 1–18, Mar. 2026, doi: 10.1080/10095020.2026.2628435.

[8] C. J. Reed *et al.*, “Scale-MAE: A Scale-Aware Masked Autoencoder for Multiscale Geospatial Representation Learning,” Sep. 22, 2023, *arXiv*: arXiv:2212.14532. doi: 10.48550/arXiv.2212.14532.

[9] O. Mañas, A. Lacoste, X. Giro-i-Nieto, D. Vazquez, and P. Rodriguez, “Seasonal Contrast: Unsupervised Pre-Training from Uncurated Remote Sensing Data,” May 03, 2021, *arXiv*: arXiv:2103.16607. doi: 10.48550/arXiv.2103.16607.

[10] X. Guo *et al.*, “SkySense: A Multi-Modal Remote Sensing Foundation Model Towards Universal Interpretation for Earth Observation Imagery,” Mar. 22, 2024, *arXiv*: arXiv:2312.10115. doi: 10.48550/arXiv.2312.10115.

[11] T. Wang and P. Isola, “Understanding Contrastive Representation Learning through Alignment and Uniformity on the Hypersphere,” Aug. 15, 2022, *arXiv*: arXiv:2005.10242. doi: 10.48550/arXiv.2005.10242.

[12] K. He, X. Chen, S. Xie, Y. Li, P. Dollár, and R. Girshick, “Masked Autoencoders Are Scalable Vision Learners,” Dec. 19, 2021, *arXiv*: arXiv:2111.06377. doi: 10.48550/arXiv.2111.06377.

[13] R. Caye Daudt, B. Le Saux, and A. Boulch, “Fully Convolutional Siamese Networks for Change Detection,” in *2018 25th IEEE International Conference on Image Processing (ICIP)*, Oct. 2018, pp. 4063–4067. doi: 10.1109/ICIP.2018.8451652.

[14] S. Fang, K. Li, J. Shao, and Z. Li, “SNUNet-CD: A Densely Connected Siamese Network for Change Detection of VHR Images,” *IEEE Geoscience and Remote Sensing Letters*, vol. 19, pp. 1–5, 2022, doi: 10.1109/LGRS.2021.3056416.

[15] A. Dosovitskiy *et al.*, “An Image is Worth 16x16 Words: Transformers for Image Recognition at Scale,” Jun. 03, 2021, *arXiv*: arXiv:2010.11929. doi: 10.48550/arXiv.2010.11929.

[16] H. Chen, Z. Qi, and Z. Shi, “Remote Sensing Image Change Detection With Transformers,” *IEEE Transactions*

*on Geoscience and Remote Sensing*, vol. 60, pp. 1–14, 2022, doi: 10.1109/TGRS.2021.3095166.
[17] W. G. C. Bandara and V. M. Patel, "A Transformer-Based Siamese Network for Change Detection," in *IGARSS 2022 - 2022 IEEE International Geoscience and Remote Sensing Symposium*, Jul. 2022, pp. 207–210. doi: 10.1109/IGARSS46834.2022.9883686.
[18] C. Wu, L. Zhang, and B. Du, "Kernel Slow Feature Analysis for Scene Change Detection," *IEEE Trans. Geosci. Remote Sensing*, vol. 55, no. 4, pp. 2367–2384, Apr. 2017, doi: 10.1109/TGRS.2016.2642125.
[19] T. Suzuki, S. Shirakabe, Y. Miyashita, A. Nakamura, Y. Satoh, and H. Kataoka, "Semantic Change Detection with Hypermaps," Mar. 16, 2017, *arXiv*: arXiv:1604.07513. doi: 10.48550/arXiv.1604.07513.
[20] A. Varghese, J. Gubbi, A. Ramaswamy, and P. Balamuralidhar, "ChangeNet: A Deep Learning Architecture for Visual Change Detection," presented at the Proceedings of the European Conference on Computer Vision (ECCV) Workshops, 2018, pp. 0–0. Accessed: Feb. 05, 2026. [Online]. Available: https://openaccess.thecvf.com/content_eccv_2018_workshops/w7/html/Varghese_ChangeNet_A_Deep_Learning_Architecture_for_Visual_Change_Detection_ECCVW_2018_paper.html
[21] R. Caye Daudt, B. Le Saux, A. Boulch, and Y. Gousseau, "Multitask learning for large-scale semantic change detection," *Computer Vision and Image Understanding*, vol. 187, p. 102783, Oct. 2019, doi: 10.1016/j.cviu.2019.07.003.
[22] Y. Wang, B. Du, L. Ru, C. Wu, and H. Luo, "Scene Change Detection VIA Deep Convolution Canonical Correlation Analysis Neural Network," in *IGARSS 2019 - 2019 IEEE International Geoscience and Remote Sensing Symposium*, Yokohama, Japan: IEEE, Jul. 2019, pp. 198–201. doi: 10.1109/IGARSS.2019.8898211.
[23] D. Peng, L. Bruzzone, Y. Zhang, H. Guan, and P. He, "SCDNET: A novel convolutional network for semantic change detection in high resolution optical remote sensing imagery," *International Journal of Applied Earth Observation and Geoinformation*, vol. 103, p. 102465, Dec. 2021, doi: 10.1016/j.jag.2021.102465.
[24] Y. Zhu, L. Li, K. Chen, C. Liu, F. Zhou, and Z. Shi, "Semantic-CD: Remote Sensing Image Semantic Change Detection towards Open-vocabulary Setting," Jan. 12, 2025, *arXiv*: arXiv:2501.06808. doi: 10.48550/arXiv.2501.06808.
[25] Y. Deng *et al.*, "Feature-Guided Multitask Change Detection Network," *IEEE J. Sel. Top. Appl. Earth Observations Remote Sensing*, vol. 15, pp. 9667–9679, 2022, doi: 10.1109/JSTARS.2022.3215773.
[26] S. Xiang, M. Wang, X. Jiang, G. Xie, Z. Zhang, and P. Tang, "Dual-Task Semantic Change Detection for Remote Sensing Images Using the Generative Change Field Module," *Remote Sensing*, vol. 13, no. 16, p. 3336, Aug. 2021, doi: 10.3390/rs13163336.
[27] L. Ding, J. Zhang, H. Guo, K. Zhang, B. Liu, and L. Bruzzone, "Joint Spatio-Temporal Modeling for Semantic Change Detection in Remote Sensing Images," *IEEE Trans. Geosci. Remote Sensing*, vol. 62, pp. 1–14, 2024, doi: 10.1109/TGRS.2024.3362795.
[28] J. Wang *et al.*, "MSCD-Net: From Unimodal to Multimodal Semantic Change Detection," *IEEE Trans. Geosci. Remote Sensing*, vol. 63, pp. 1–17, 2025, doi: 10.1109/TGRS.2025.3591814.
[29] Q. Shu *et al.*, "Semantic Change Detection of Roads and Bridges: A Fine-grained Dataset and Multimodal Frequency-driven Detector," Sep. 19, 2025, *arXiv*: arXiv:2505.13212. doi: 10.48550/arXiv.2505.13212.
[30] B. Wijenayake *et al.*, "Mamba-FCS: Joint Spatio-Frequency Feature Fusion, Change-Guided Attention, and SeK Loss for Enhanced Semantic Change Detection in Remote Sensing," Aug. 11, 2025, *arXiv*: arXiv:2508.08232. doi: 10.48550/arXiv.2508.08232.
[31] Y. Chen, C. Li, C. Ling, Y. Tan, and P. Wu, "FAPMNet: Flow-Aligned Prototype Memory Network for Semantic Change Detection in Remote Sensing Images," *IEEE Geoscience and Remote Sensing Letters*, vol. 23, pp. 1–5, 2026, doi: 10.1109/LGRS.2026.3652300.
[32] X. Liu *et al.*, "GSTM-SCD: Graph-enhanced spatio-temporal state space model for semantic change detection in multi-temporal remote sensing images," *ISPRS Journal of Photogrammetry and Remote Sensing*, vol. 230, pp. 73–91, Dec. 2025, doi: 10.1016/j.isprsjprs.2025.09.003.
[33] E. J. Parelius, "A Review of Deep-Learning Methods for Change Detection in Multispectral Remote Sensing Images," *Remote Sensing*, vol. 15, no. 8, p. 2092, Apr. 2023, doi: 10.3390/rs15082092.
[34] G. Cheng *et al.*, "Change Detection Methods for Remote Sensing in the Last Decade: A Comprehensive Review," *Remote Sensing*, vol. 16, no. 13, p. 2355, Jan. 2024, doi: 10.3390/rs16132355.
[35] W. Jing, K. Chi, Q. Li, and Q. Wang, "ChangeRD: A registration-integrated change detection framework for unaligned remote sensing images," *ISPRS Journal of Photogrammetry and Remote Sensing*, vol. 220, pp. 64–74, Feb. 2025, doi: 10.1016/j.isprsjprs.2024.11.019.
[36] L. Ding, H. Guo, S. Liu, L. Mou, J. Zhang, and L. Bruzzone, "Bi-Temporal Semantic Reasoning for the Semantic Change Detection in HR Remote Sensing Images," *IEEE Trans. Geosci. Remote Sensing*, vol. 60, pp. 1–14, 2022, doi: 10.1109/TGRS.2022.3154390.
[37] Z. Chen *et al.*, "Vision Transformer Adapter for Dense Predictions," Feb. 13, 2023, *arXiv*: arXiv:2205.08534. doi: 10.48550/arXiv.2205.08534.
[38] O. Ronneberger, P. Fischer, and T. Brox, "U-Net: Convolutional Networks for Biomedical Image Segmentation," in *Medical Image Computing and Computer-Assisted Intervention – MICCAI 2015*, N. Navab, J. Hornegger, W. M. Wells, and A. F. Frangi, Eds., Cham: Springer International Publishing, 2015, pp. 234–241. doi: 10.1007/978-3-319-24574-4_28.
[39] K. Yang *et al.*, "Asymmetric Siamese Networks for Semantic Change Detection in Aerial Images," *IEEE Trans. Geosci. Remote Sensing*, vol. 60, pp. 1–18, 2022, doi: 10.1109/TGRS.2021.3113912.
[40] P. Yuan, Q. Zhao, X. Zhao, X. Wang, X. Long, and Y. Zheng, "A transformer-based Siamese network and an

open optical dataset for semantic change detection of remote sensing images," *International Journal of Digital Earth*, vol. 15, no. 1, pp. 1506–1525, Dec. 2022, doi: 10.1080/17538947.2022.2111470.
[41] B. Rolih, M. Fučka, F. Wolf, and L. Č. Zajc, "Be the Change You Want to See: Revisiting Remote Sensing Change Detection Practices," *IEEE Transactions on Geoscience and Remote Sensing*, vol. 63, pp. 1–11, 2025, doi: 10.1109/TGRS.2025.3585342.
[42] K. He, X. Zhang, S. Ren, and J. Sun, "Deep Residual Learning for Image Recognition," presented at the Proceedings of the IEEE Conference on Computer Vision and Pattern Recognition, 2016, pp. 770–778. Accessed: Feb. 05, 2026. [Online]. Available: https://openaccess.thecvf.com/content_cvpr_2016/html/He_Deep_Residual_Learning_CVPR_2016_paper.html
[43] Z. Liu *et al.*, "Swin Transformer: Hierarchical Vision Transformer Using Shifted Windows," presented at the Proceedings of the IEEE/CVF International Conference on Computer Vision, 2021, pp. 10012–10022. Accessed: Feb. 05, 2026. [Online]. Available: https://openaccess.thecvf.com/content/ICCV2021/html/Liu_Swin_Transformer_Hierarchical_Vision_Transformer_Using_Shifted_Windows_ICCV_2021_paper
[44] Y. Cong *et al.*, "SatMAE: Pre-training Transformers for Temporal and Multi-Spectral Satellite Imagery".
[45] Y. Liu *et al.*, "VMamba: Visual State Space Model," *Advances in Neural Information Processing Systems*, vol. 37, pp. 103031–103063, Dec. 2024, doi: 10.52202/079017-3273.

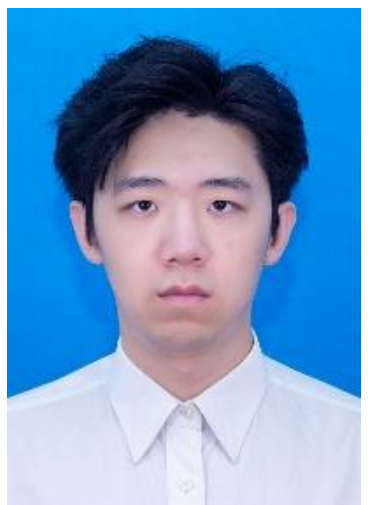

**Hengtong Shen** received the M.E. degree in photogrammetry and remote sensing from the Chinese Academy of Surveying and Mapping in 2025. He is currently pursuing a Ph.D. degree at Wuhan University.

His research interests include intelligent remote sensing interpretation, self-supervised learning for remote sensing, and deep learning.

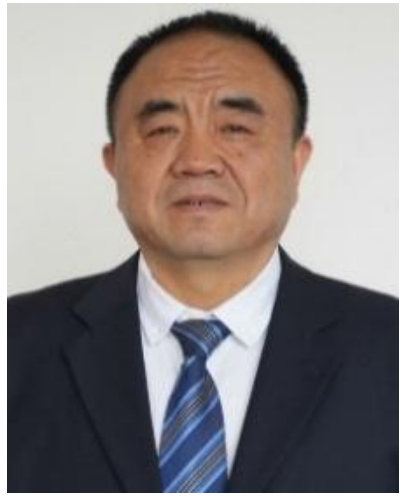

**Li Yan** received the B.S., M.S., and Ph.D. degrees in photogrammetry and remote sensing from Wuhan University, Wuhan, China, in 1989, 1992, and 1999, respectively.

He is currently a Luojia Distinguished Professor with the School of Geodesy and Geomatics, Wuhan University. His research interests include 3-D reconstruction and measurement, real-time mobile mapping and surveying, intelligent remote sensing, and precise image measurement.

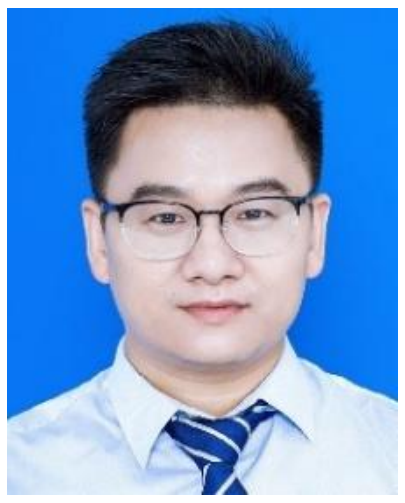

**Hong Xie** received the B.S., M.S., and Ph.D. degrees in photogrammetry and remote sensing from Wuhan University, Wuhan, China, in 2007, 2009, and 2013, respectively. He is currently an Associate Professor with the School of Geodesy and Geomatics, Wuhan University.

His research interests include target detection based on image deep learning, point cloud data quality improvement, point cloud information extraction and model reconstruction, mobile mapping, and surveying.

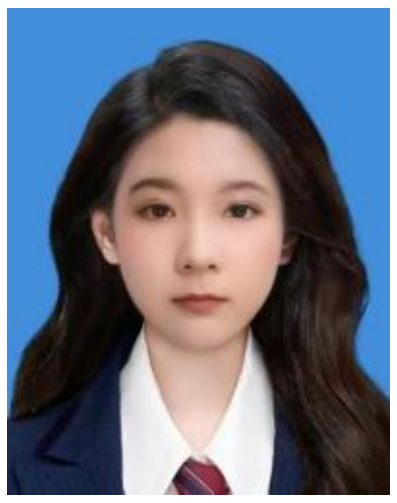

**Yaxuan Wei** received the M.E. degree in Surveying and Mapping from Beijing University of Civil Engineering and Architecture in 2025.

Her research interests include 3D change detection, semantic segmentation, and deep learning.

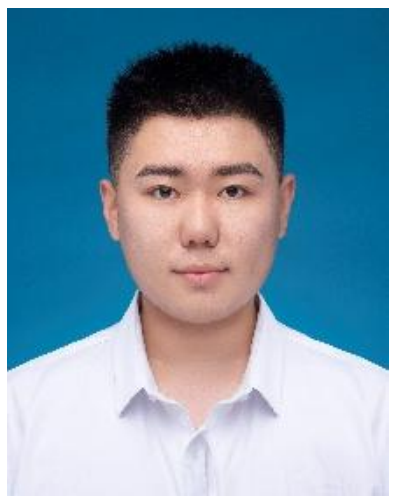

**Xinhao Li** received the M.E. degree in Surveying and Mapping from China University of Petroleum (East China) in 2025.

His research interests include 3D semantic segmentation, intelligent remote sensing interpretation, and deep learning.

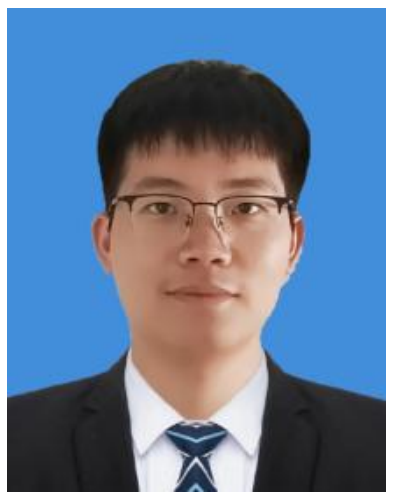

**Wenfei Shen** received the M.E. degree in Surveying and Mapping from Beijing University of Civil Engineering and Architecture in 2025.

His research interests include 3D gaussian splatting reconstruction, temporal reconstruction, and deep learning.

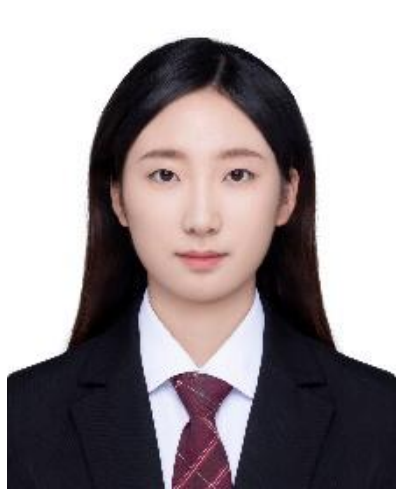

**Peixian Lv** received the B.E. degree in surveying and mapping engineering from Nanjing Normal University in 2025. She is currently pursuing a M.E. degree at Wuhan University.

Her research interests focus on 3D change detection.

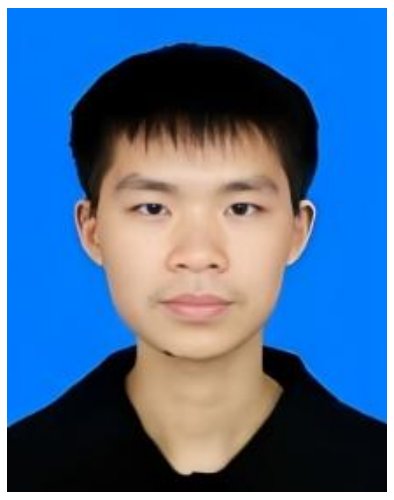

**Fei Tan** received his B.E. degree in remote sensing science and technology from Southwest Jiaotong University in 2024. He is currently pursuing a M.E. degree at Wuhan University.

His research interests include change detection in urban scenes and point cloud deep learning.